\documentclass{article}

    \usepackage[final]{neurips_2022}

\usepackage[utf8]{inputenc} % allow utf-8 input
\usepackage[T1]{fontenc}    % use 8-bit T1 fonts
\usepackage{hyperref}       % hyperlinks
\usepackage{url}            % simple URL typesetting
\usepackage{booktabs}       % professional-quality tables
\usepackage{amsfonts}       % blackboard math symbols
\usepackage{nicefrac}       % compact symbols for 1/2, etc.
\usepackage{microtype}      % microtypography
\usepackage{xcolor}         % colors

\usepackage{nicefrac}       % compact symbols for 1/2, etc.
\usepackage{microtype}      % microtypography
\usepackage{algorithm}
\usepackage{algorithmic}
\usepackage{amsmath}
\usepackage{graphicx}
\usepackage{caption}
\usepackage{subcaption}
\usepackage{xcolor}

\usepackage{pifont}% http://ctan.org/pkg/pifont
\newcommand{\cmark}{\ding{51}}%
\newcommand{\xmark}{\ding{55}}%

\usepackage{physics}
\usepackage{multirow}

\usepackage{amsthm}
\usepackage[normalem]{ulem}

\graphicspath{{./},{../figures/},{./figures/}}

\newtheorem{prop}{Proposition}

\usepackage{soul}

\newcommand{\sL}{\bar{{\mathbf L}}^q}
\newcommand{\nsL}{\bar{{\mathbf L}}^q_{\mathrm{n}}}

\newcommand{\ba}{{\mathbf a}}
\newcommand{\bg}{{\mathbf g}}
\newcommand{\bh}{{\mathbf h}}
\newcommand{\bx}{{\mathbf x}}

\newcommand{\by}{{\mathbf y}}

\newcommand{\bs}{{\mathbf s}}
\newcommand{\bu}{{\mathbf u}}

\newcommand{\bA}{{\mathbf A}}
\newcommand{\bD}{{\mathbf D}}
\newcommand{\bG}{{\mathbf G}}
\newcommand{\bH}{{\mathbf H}}
\newcommand{\bI}{{\mathbf I}}

\newcommand{\bL}{{\mathbf L}}

\newcommand{\bP}{{\mathbf P}}

\newcommand{\bW}{{\mathbf W}}

\newcommand{\bX}{{\mathbf X}}

\newcommand{\bPhi}{{\boldsymbol \Phi}}
\newcommand{\bLambda}{{\boldsymbol \Lambda}}

\newcommand{\bTheta}{{\boldsymbol \Theta}}

\newcommand{\btheta}{{\boldsymbol \theta}}

\newcommand{\bB}{{\mathbf B}}

\newcommand{\cE}{{\mathcal E}}

\newcommand{\cG}{{\mathcal G}}

\newcommand{\cL}{{\mathcal L}}
\newcommand{\cN}{{\mathcal N}}

\newcommand{\cS}{{\mathcal S}}
\newcommand{\cT}{{\mathcal T}}

\newcommand{\cV}{{\mathcal V}}

\def\bU{{\bf U}}

\def\minwrt[#1]{\underset{#1}{\text{minimize}}}
\def\maxwrt[#1]{\underset{#1}{\text{max}}}

\newcommand{\singh}[1]{\textcolor{black}{#1}}

\title{Signed Graph Neural Networks: A Frequency Perspective}

\author{%
  Rahul Singh\\
  Georgia Tech, Atlanta, GA \\
  \texttt{rasingh@gatech.edu} \\
   \And
   Yongxin Chen  \\
   Georgia Tech, Atlanta, GA \\
   \texttt{yongchen@gatech.edu} \\
}

\begin{document}

\maketitle
%%%%%%%%%%%%%%%%%%%%%%%%%%%%%%%%%
\begin{abstract}
Graph convolutional networks (GCNs) and its variants are designed for unsigned graphs containing only positive links. Many existing GCNs have been derived from the spectral domain analysis of signals lying over (unsigned) graphs and in each convolution layer they perform low-pass filtering of the input features followed by a learnable linear transformation. Their extension to signed graphs with positive as well as negative links imposes multiple issues including computational irregularities and ambiguous frequency interpretation, making the design of computationally efficient low pass filters challenging. In this paper, we address these issues via spectral analysis of signed graphs and propose two different signed graph neural networks, one keeps only low-frequency information and one also retains high-frequency information. We further introduce magnetic signed Laplacian and use its eigendecomposition for spectral analysis of directed signed graphs. We test our methods for node classification and link sign prediction tasks on signed graphs and achieve state-of-the-art performances.
\end{abstract}
%%%%%%%%%%%%%%%%%%%%%%%%%%%%%%%%%%%%
%%%%%%%%%%%%%%%%%%%%%%%%%%%%%%%%%%%%%%%%%%%%
\section{Introduction} \label{sec:Intro}

% \begin{itemize}
%     \item Signed graphs are important
%     \item The magnetic Laplacian is not directly applicable to signed graphs
%     \item Existing methods on signed graphs
%     \item we propose
% \end{itemize}

Graph neural networks (GNNs) learn powerful node representations by capturing local graph structure and feature information~\citep{MaTan21,WuPanChe21}. The existing GNN architectures have focused almost exclusively on graphs with nonnegative edges, which encode some kind of similarity relation between the incident nodes. In contrast, negative edges are often useful to model dissimilarity relations~\citep{KumSpeSub16,DitMat20}: for instance, in social networks, users may have common/opposite political views, trust/distrust one another’s recommendations, or like/dislike each other. Such dissimilarity relations can be modeled using signed graphs by allowing the edges to take both positive or negative values. 
% Another example of signed graph data is gene regulatory networks~\citep{LiKexMar21,KarSahAvi21} where the sign of regulation links represents an activation (positive) or a repression (negative).
In this paper, we are interested in graph neural networks designs for signed graphs. 
%The existing GNNs cannot be directly applied to signed graphs due to its underlying similarity relationship assumption. 

There are two major lines of research that 
% Multiple efforts have been made to 
consider signed links in the process of learning the node embeddings. 
On one hand, the network embedding methods such as SIDE~\citep{KimParLee18} and SLF~\citep{XuHuWu19} learn node representations by optimizing an unsupervised loss that primarily aims to locate node embeddings closer to each other if they are connected by positive links and vice-versa for nodes connected by negative links. On the other hand, SGCN~\citep{DerMaTan18}, SiGAT~\citep{HuaSheHou19}, and SNEA~\citep{LiTiaZha20} adopt GNN based models to learn node embeddings in a task-specific end-to-end manner. These GNN based models are based on structural balance theory for signed graphs~\citep{Hei46,CarHar56}. 

% Recently, theory for spectral domain analysis of signed graphs has been developed~\citep{DitMat20} providing frequency interpretation for features lying on the signed graphs that we use in our GNN designs. 

% Current GNNs can be classified into mainly two categories~\citep{WuPanChe21}: spectral domain designs and spatial domain designs. Spectral domain GNN designs including spectral-GNN~\citep{BruZarSzl14} and ChebNet~\citep{DefBreXav16} are based on graph signal processing with graph convolutions defined in the frequency domain of the underlying graph. On the other hand, the spatial domain GNN designs work with the graph convolutions defined in the graph spatial domain.  
In this work, we propose an alternative solution to GNNs for signed graphs from a frequency perspective via spectral domain analysis.
Recall that the spectral domain analysis of unsigned graphs has been widely used to develop GNN architectures. Many well-known GNNs including spectral-GNN~\citep{BruZarSzl14}, ChebNet~\citep{DefBreXav16}, GCN~\citep{KipWel17}, AGCN~\citep{LiWanZhu18}, and FAGCN~\citep{BoXiaChu21} rely on spectral domain analysis. These designs are based on the frequency interpretation derived from the eigendecomposition of the normalized unsigned graph Laplacian. However, direct application of the existing spectral domain GNN designs  to signed graphs is problematic, mainly due to (i) possible zero diagonal entries in the degree matrix making the normalization of the Laplacian prohibitive and (ii) possible negative eigenvalues of the graph Laplacian, making the frequency ordering somewhat ambiguous, i.e., whether the smallest negative, positive, or absolute value, eigenvalues should be used as low frequency~\citep{Kny17}.

To address these issues, we turn to signed graph signal processing~\citep{DitMat20} which provides frequency interpretation for features lying on the signed graphs, and propose spectral domain signed GNNs based on it. Specifically, we propose two different GNN designs for signed graphs: Spectral-SGCN-I and Spectral-SGCN-II. The former considers fixed low-pass filter keeping only low-frequency information during aggregation process, whereas the later is based on attention mechanism retaining low as well as high-frequency information. Extending these methods to directed signed graphs is another challenge. For handling directed signed graph, we further introduce spectral methods for directed signed graphs. We evaluate the performance of our methods on node classification and link sign prediction tasks on signed graphs. 
Our contributions are summarized below.
\begin{itemize}
    \item We present a principled approach to designing graph neural networks for signed graphs based on the spectral domain analysis over signed graphs.  
    \item We instantiate our approach with two graph neural network architectures for signed graphs, one behaves like a low-pass filter and one also retains high-frequency information.
    \item  We introduce signed magnetic Laplacian (see Table \ref{table:laplacians}) for spectral analysis of directed signed graphs and utilize it in feature aggregation process.
    \item We evaluate our method through extensive evaluations on node classification as well as link sign prediction tasks for signed graphs and achieve state-of-the-art performances.
\end{itemize}
%%%%%%%%%%%%%%%%%%%%%%%%%%%%%%%%%%%%%%%%%%%
%--------------------------------------
\begin{table*}[ht]
\scriptsize
\caption{Different Laplacians and their applicability.}
\vspace{-0.4cm}
\begin{center}
\begin{tabular}{|c|c|c|c|c|}
\hline
   & Unsigned & Directed & Signed &  Directed Signed  \\
\hline\hline
  Laplacian $\bL$ & \cmark & \xmark  &  \xmark & \xmark \\
  Magnetic Laplacian $\bL^q$ & \cmark & \cmark  &  \xmark & \xmark \\
  Signed Laplacian $\bar{\bL}$ & \cmark & \xmark  &  \cmark & \xmark \\
  Signed Magnetic Laplacian $\bar{\bL}^q$ & \cmark & \cmark  &  \cmark & \cmark \\
\hline
\end{tabular}
\end{center}
\label{table:laplacians}
\end{table*}
%_-----------------------------------------
% There exist many different methods to learn node representations (embeddings) on signed graphs.
\textbf{Related Work:}  Most of the existing methods for signed graphs are derived from balance theory and can be classified into two categories: unsupervised network embeddings and GNN based methods. Unsupervised network embedding methods including SiNE~\citep{WanTanAgg17}, SIDE~\citep{KimParLee18}, SIGNet~\citep{IslPraRam18}, SLF~\citep{XuHuWu19}, and ASiNE ~\citep{LeeSeoHan20} learn node representations in an unsupervised manner. These node embeddings are then used for task in hand separately. GNN based methods including signed graph convolutional network (SGCN)~\citep{DerMaTan18}, signed graph attention network (SiGAT)~\citep{HuaSheHou19}, signed network embedding based on attention (SNEA)~\citep{LiTiaZha20}, and group signed graph neural network~(GS-GNN)~\citep{LiuZhaCui21} are jointly trained to learn node embeddings along with the task in hand in an end-to-end manner. SGCN is the state-of-the-art signed GNN model considering balanced and
unbalanced paths motivated from the balance theory to aggregate local graph information with fixed coefficients. Different from these, our proposed methods are based on spectral domain analysis of signed graphs. \singh{SDGNN~\citep{HuaSheHou21} is a recent work applicable to signed directed graphs based on balance and status theory. SSSNET~\citep{HeReiWan22} is another GNN based work with a focus on clustering of signed graphs. For balanced graphs, the eigenvectors of the signed Laplacian follow certain properties as analyzed in \citep{DitMat20}. However, it is difficult to relate the signed spectral analysis with balance theory in case of unbalanced graphs. It is an interesting problem to explore the relationship between balance theory and spectral analysis for unbalanced graphs, which is out of scope of this work.} 

% SNEA further extended SGCN to incorporate learnable attention coefficients for aggregating balanced and unbalanced paths. SiGAT is a motif-based GNN model to learn the node representation inspired by GAT~\citep{VelGuiAra18}.
% Different from these works, we make use of frequency analysis of signed graphs via eigendecomposition of the signed Laplacian for new \singh{\textit{interpretable}} signed GNN designs. Note that the signed Laplacian has been used in signed graph learning~\citep{MatDit20} and its eigendecomposition has been utilized for community detection and spectral clustering~\citep{KunSchLom10,Kny17} tasks in signed graphs. However in this work, we use signed Laplacian in representation learning for signed graphs.

%%%%%%%%%%%%%%%%%%%%%%%%%%%%%%%%%%%%%%%%%%%%
\section{Preliminaries} \label{sec:Preliminaries}
%%%%%%%%%%%%%%%%%%%%%%%%%%%%%%%%%%%%%%%%%%%%
Let $\cG = (\cV,\cE^+ \cup \cE^-)$ be a directed signed graph, where $\cV$ is the set of $N$ number of nodes, $\cE^+$ is the set of directed positive edges, and $\cE^-$ is the set of directed negative edges. The adjacency matrix of the graph is denoted as $\bA\in \mathbb{R}^{N\times N}$ and has entries from $\{+1,-1,0 \}$\singh{\footnote{For weighted signed graphs, the weighed adjacency matrix can be used taking any real value as edge weights. Our analysis and GNN models are applicable to general weighted signed graphs.}}. If $(i,j)$ is not an edge of the graph, then the corresponding entry is $\bA(i,j) = 0$. $\bA(i,j) = +1$ denotes a positive edge from node $i$ to $j$, whereas $\bA(i,j) = -1$ denotes a negative edge from node $i$ to $j$. Denote $\cN_i^+$ as the set of neighbors connected to node $i$ via positive edges and $\cN_i^-$ as the set of neighbors connected to node $i$ via negative edges. In addition, a feature matrix $\bX \in \mathbb{R}^{N \times F} $ is utilized to describe nodes properties (input features), with $\bx_i \in \mathbb{R}^N$ (column of $\bX$) representing the $i^{th}$ feature channel of $\bX$ and $F$ denotes the total number of feature channels.

%============================
\subsection{Traditional GSP and Spectral Domain GNN Designs}
\label{subsec:traditional_GNNs}
Popular spectral domain designs of graph neural networks including ChebNet~\citep{DefBreXav16}, GCN~\citep{KipWel17} and their further improvements such as AGCN~\citep{LiWanZhu18}, Simplified GCN~\citep{WuSouZha19}, and FAGCN~\citep{BoXiaChu21} are based on the spectral analysis of signals (features) defined on an unsigned graph. The spectral analysis of graph signals has been studied under the umbrella of the graph signal processing (GSP) framework~\citep{ShuNarFro13,OrtFroKov18}. The graph Fourier (spectral) analysis relies on the spectral decomposition of graph Laplacians. The traditional combinatorial graph Laplacian is defined as $\bL = \bD - \bA$, with $\bD = \mathrm{diag}\{ d_1,d_2,\ldots, d_N \}$ and $d_i  = \sum_j A_{ij}$; its normalized version is $\bL_{\mathrm{n}} = \bD^{-1/2} \bL \bD^{-1/2}$. Based on the eigendecomposition of the graph Laplacian $\bL = \bU \bLambda \bU^T$, where $\bU \in \mathbb{R}^{N\times N}$ comprises of orthonormal eigenvectors and $\bLambda = \mathrm{diag}\{\lambda_1, \ldots, \lambda_N\}$ is a diagonal matrix of eigenvalues, the graph Fourier transform is defined with eigenvectors of the graph Laplacian being the graph Fourier modes (harmonics) and the corresponding eigenvalues being the graph frequencies~\citep{ShuNarFro13}. Assuming $\lambda_1 \leq \lambda_2 \leq \ldots \leq \lambda_N$, $\lambda_1$ corresponds to the lowest (zero) frequency and $\lambda_N$ corresponds to the highest frequency of the graph. For the case of normalized Laplacian $\bL_{\mathrm{n}}$, all the graph frequencies lie in the range $[0,2]$~\citep{ShuNarFro13}, with $\lambda_1 = 0$.  

Let $\bx \in \mathbb{R}^{N}$ be a single-channel input signal on the graph, then the graph Fourier transform and the inverse Fourier transform are defined as $\hat{\bx} =  \bU^T \bx$ and $\bx = \bU \hat{\bx}$, respectively.  Graph convolution of the input graph signal $\bx$ with a filter $\bg$ is $\bx * \bg := \bU \hat{\bG} \bU^T \bx$, where $\hat{\bG} := \mathrm{diag} (\hat\bg)=\mathrm{diag}\{\hat{g}_1,  \ldots, \hat{g}_N\}$. For computational efficiency, the filter coefficients $\hat{g}_1,  \ldots, \hat{g}_N$ can be approximated via $K^{th}$ order polynomials of the graph frequencies ($K<<N$), i.e., $\hat{g}(\lambda_j) = \sum_{i=0}^{K} \theta_i \lambda_j^i$ with $\btheta \in \mathbb{R}^{K+1}$ being the (polynomial) filter coefficients. 
% Then the graph convolution with such polynomial filter takes the form
% \begin{equation}
%     \bx * \bg \approx \bU \left( \sum_{i=0}^K \theta_i \bLambda^i \right) \bU^T \bx = \sum_{i=0}^K \theta_i \bL_{\mathrm{n}} ^i \bx.
% \end{equation}
% Note that the polynomial filters allows us to write graph convolution in spatial domain via polynomial in the graph Laplacian. ChebNet~\citep{DefBreXav16} proposed to approximate the graph convolution via $K^{th}$ order Chebyshev polynomials.
In GCN, \citep{KipWel17} simplified the graph convolution by assuming first order polynomial filter ($K=1$) with $\theta_0 = 2 \theta$ and $\theta_1 = - \theta$, and thereby reducing the graph convolution to
\begin{equation}
    \bx * \bg \approx \theta~ (2 \bI - \bL_{\mathrm{n}})~ \bx = \theta ~(\bI + \bD^{-1/2} \bA \bD^{-1/2})~ \bx.
\end{equation}

As a different interpretation, the above can also be viewed as a combination of two operations: (i) Feature aggregation via term ($\bI + \bD^{-1/2} \bA \bD^{-1/2}$) and (ii) Feature transformation via learnable parameter $\theta$. Note that the feature aggregation operation corresponds to low pass filtering since the spectral response of the spatial filter $2 \bI - \bL_{\mathrm{n}}$ is $\hat{g}(\lambda) = 2 - \lambda$ which amplifies low-frequency signal ($\lambda\approx 0$) and restrains high-frequency signal ($\lambda \approx 2$). In its final design, for numerical stability, the feature aggregation operation in GCN is modified by adding self-loops for each node and as a result the modified graph convolution takes the form~\citep{KipWel17,WuSouZha19}
\begin{equation}\label{eq:renormalization}
     \bx * \bg \approx \theta ~(\Tilde{\bD}^{-1/2} \Tilde{\bA} \Tilde{\bD}^{-1/2})~ \bx, 
\end{equation}
where $\Tilde{\bA} = \bA + \bI$ and $\Tilde{\bD} = \bD + \bI$. When generalized to multi-channel input $\bX$, the output of the $\ell^{th}$ layer of the GCN reads
\begin{equation}
    \bH^{(\ell)} = \sigma ( \bP~ \bH^{(\ell - 1)}~ \bTheta ^{(\ell)} ) , \quad \quad \bH^{(0)} = \bX,
\end{equation}
where $\bP = \Tilde{\bD}^{-1/2} \Tilde{\bA} \Tilde{\bD}^{-1/2}$ is the low-pass feature aggregation filter, $\bTheta^{(\ell)}$ is a learnable transformation matrix, and $\sigma$ is non-linearity such as ReLU. 
% Stacking $L$ number of GCN layers, we can effectively aggregate features from nodes that are $L$-hops away from each node in the graph. 

% When using the traditional definition of the Laplacian matrix for the case of signed graphs, it imposes the following issues.
% \begin{itemize}
%     \item 
% \end{itemize}
\subsection{Issues with Signed Graphs}
\label{subsec:issue_signed}
In each GCN layer, the feature aggregation operation corresponds to low pass filtering with the filter being first order polynomial in $\bL_{\mathrm{n}}$. However for signed graphs, the inverse of the degree matrix $\bD$ (or $\Tilde{\bD}$) becomes problematic since the degree values might be zero or negative values for some nodes and as a consequence the normalized Laplacian $\bL_{\mathrm{n}}$ is not well defined. 
% This can be clearly seen as a toy example graph shown in Figure~\ref{fig:toy1a} with $\bD = \mathrm{diag}\{-1, 0, -2, -1 \}$ which is not invertible.
Since the normalized Laplacian is not well defined, one is tempted to interpret the aggregation operator as a low pass filter based on unnormalized Laplacian $\bL$. This again poses difficulty in frequency ordering as the eigenvalues of the Laplacian $\bL$ can take negative values for signed graphs. The graph frequencies (Laplacian eigenvalues) are ordered based on the total variation (TV) of the corresponding eigenvectors on the graph~\citep{SanMou14,OrtFroKov18}. TV quantifies global smoothness (or variation) of a graph signal. For unsigned graphs, the quadratic form $\bx^T \bL \bx$ is often used as TV of signal $\bx$ on the graph~\citep{ShuNarFro13}. However for signed graphs, the quadratic form $\bx^T \bL \bx$ may take negative values and thereby invalidating its use as TV. Another definition of TV of a graph signal $\bx$ is~\citep{SinChaMan16} $\mathrm{TV(\bx)} = || \bL \bx||_1$
and it can be shown that $ \mathrm{TV(\bu_i)} > \mathrm{TV(\bu_j)},~ \mathrm{if}~ |\lambda_i| > | \lambda_j|$. Thus the eigenvalues with smaller absolute values act as low frequencies and vice-versa. 
%*************************
\begin{figure}[ht]
\vspace{-0.4cm}
\centering
\begin{subfigure}[t]{0.15\textwidth}
\centering	
\includegraphics[scale=0.7]{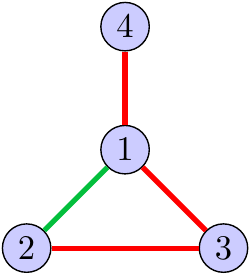}
\vspace{-0.2cm}
\caption{}
\label{fig:toy1a}
\end{subfigure}
\begin{subfigure}[t]{0.4\textwidth}
\centering
\includegraphics[scale=0.5]{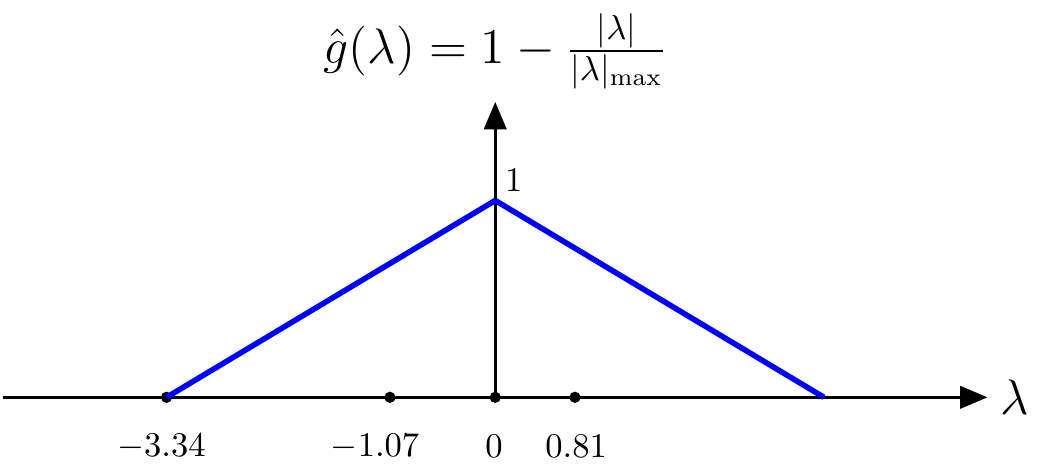}
\vspace{-0.2cm}
\caption{}
\label{fig:toy1b}
\end{subfigure}
\begin{subfigure}[t]{0.2\textwidth}
\centering	
\includegraphics[scale=0.6]{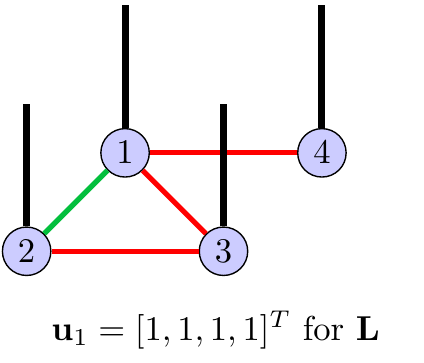}
\vspace{-0.2cm}
\caption{}
\label{fig:toy1c}
\end{subfigure}
\begin{subfigure}[t]{0.2\textwidth}
\centering
\includegraphics[scale=0.6]{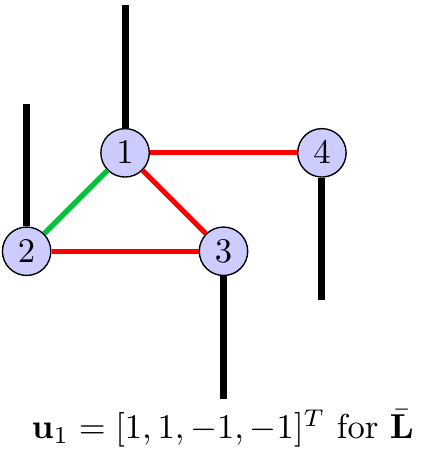}
\vspace{-0.2cm}
\caption{}
\label{fig:toy1d}
\end{subfigure}
\vspace{-0.2cm}
\caption{(a) A toy signed graph (Red color edges represent negative links and Green color edges represent positive links), (b) corresponding low-pass filter based on the unnormalized Laplacian, (c) Eigenvector corresponding to the lowest frequency based on the unsigned Laplacian, and (d) Eigenvector corresponding to the lowest frequency based on the signed Laplacian.}
\label{fig:toy1}
\end{figure}
% *************************
% \begin{figure}[ht]
% \centering
% \begin{subfigure}[t]{0.45\textwidth}
% \centering	
% \includegraphics[scale=0.6]{./figures/eig_unsigned.pdf}
% % \caption{An example directed graph.}
% \end{subfigure}
% \begin{subfigure}[t]{0.45\textwidth}
% \centering
% \includegraphics[scale=0.6]{./figures/eig_signed.pdf}
% % \caption{An example directed graph.}
% \end{subfigure}
% \caption{Eigenvectors corresponding to the lowest frequencies based on the unsigned Laplacian (left) and signed Laplacian (right).}
% \label{fig:eigen_vectors}
% \end{figure}
% %*************************

Although one can order the signed frequencies based on the absolute eigenvalues of the graph Laplacian $\bL$, one needs new designs of low pass filters to be used as aggregation operator. For example, a low pass filter for the toy graph in Figure~\ref{fig:toy1a} with frequency response $ \hat{g}(\lambda) =  1- \frac{|\lambda|}{|\lambda|_{\mathrm{max}}}$ is shown in Figure~\ref{fig:toy1b} where $|\lambda|_{\mathrm{max}}$ is the maximum absolute frequency of the underlying graph. This filter design has certain drawbacks since it requires computation of $|\lambda|_{\mathrm{max}}$ and cannot be directly realized as a first order polynomial (in the graph Laplacian) in the spatial domain 
as the latter corresponds to a straight line. One can go for higher order filters with additional computational cost. 

Moreover, the frequency interpretation also becomes ambiguous for signed graphs. Since the zero eigenvalue being the lowest frequency and the corresponding eigenvector being a constant signal on the graph (as shown in Figure~\ref{fig:toy1c}), it is intuitive only under similarity assumption (i.e. positive edges). For example, foes/enemies (users connected via negative links) having similar opinions (traits or features) suggests a high amount of variation and friends (users connected via positive links) having similar opinions constitutes small variation. However, this intuition is violated when using unsigned Laplacian: lowest frequency eigenvector values at nodes 1 and 3 connected via negative links in Figure~\ref{fig:toy1c} have same values. 

% %===============================
% \subsection{Signed Laplacian}\label{subsec:signed_Laplacian}
% The signed Laplacian matrix~\citep{KunSchLom10,DitMat20} is defined as
% \begin{equation}\label{eq:signedL}
%     \Bar{\bL} = \Bar{\bD} - \bA,
% \end{equation}
% where $\Bar{\bD} = \mathrm{diag} \{\Bar{d}_1, \ldots, \Bar{d}_N \}$ is a diagonal matrix with $\Bar{d}_{i} = \sum_{j} |A_{ij}|$. In \citep{DitMat20} the authors formalized the spectral domain analysis of the signals over signed graph via eigendecomposition of the signed Laplacian matrix with the eigenvalues being the signed graph frequencies and the corresponding eigenvectors being the signed graph harmonics. The normalized signed Laplacian matrix is
% \begin{equation}
%     \Bar{\bL}_{\mathrm{n}} = \Bar{\bD}^{-1/2} \Bar{\bL} \Bar{\bD}^{-1/2} = \bI - \Bar{\bD}^{-1/2} \bA \Bar{\bD}^{-1/2}.
% \end{equation}
% The eigenvalues of the normalized signed Laplacian lie in the range $[0,2]$~\citep{SinChe22} with smaller eigenvalues corresponding to low frequencies, and vice-versa. \singh{The signed Laplacian cannot handle directed edges in signed graphs. }
%%%%%%%%%%%%%%%%%%%%%%%%%%%%%%%%%%%%%%%%%%%
\section{Proposed Method}
\label{sec:proposed}
%----------------------------------
In this section we present our spectral domain analysis approach to graph neural networks for signed graphs. We then instantiate the approach to two specific network designs. 
%====================================
\subsection{Spectral Domain Analysis of Signed Graphs}
%As we have seen in Section \ref{subsec:issue_signed}, the standard graph Laplacian does not lead to a meaningful frequency domain interpretation for signed graphs.
To address the issues mentioned in Section \ref{subsec:issue_signed} while retaining the frequency interpretation of the aggregation operation, we turn to the signed graph signal processing~\citep{DitMat20}. Instead of the standard graph Laplacian, we consider the signed Laplacian matrix~\citep{KunSchLom10,DitMat20} $\Bar{\bL} = \Bar{\bD} - \bA$, where $\Bar{\bD} = \mathrm{diag} \{\Bar{d}_1, \ldots, \Bar{d}_N \}$ is a diagonal matrix with $\Bar{d}_{i} = \sum_{j} |A_{ij}|$. In \citep{DitMat20} the authors formalized the spectral domain analysis of the signals over signed graph via eigendecomposition of the signed Laplacian matrix with the eigenvalues being the signed graph frequencies and the corresponding eigenvectors being the signed graph harmonics. 

More precisely, we consider the normalized signed Laplacian matrix
\begin{equation}\label{eq:normalized_signed_Laplacian}
    \Bar{\bL}_{\mathrm{n}} = \Bar{\bD}^{-1/2} \Bar{\bL} \Bar{\bD}^{-1/2} = \bI - \Bar{\bD}^{-1/2} \bA \Bar{\bD}^{-1/2}.
\end{equation}
The eigenvalues of the normalized signed Laplacian lie in the range $[0,2]$ with smaller eigenvalues corresponding to low frequencies, and vice-versa. This frequency ordering directly follows from using quadratic Laplacian form 
\singh{
\begin{equation}
   \mathrm{TV(\bx)} =  \bx^T \Bar{\bL}_{\mathrm{n}} \bx
\end{equation}
}
as the definition of TV on signed graphs. Using the eigenvectors of the signed Laplacian as graph harmonics provides natural frequency interpretation for signed graphs. The eigenvector corresponding to the lowest frequency of a signed graph is shown in Figure~\ref{fig:toy1d}. It can be seen that the nodes connected via negative links (foes) have opposite values and nodes connected via positive links (friends) have similar values thereby exhibiting small amount of variation; this phenomenon is intuitive for being a low frequency signal. 

% \begin{itemize}
%     \item Use $\Bar{\bA} = \bI + \bA$ added self loop. (later Renormalization trick)
% \end{itemize}

Our approach to signed graph neural networks naturally follows by redefining graph convolution based on the normalized signed Laplacian~\eqref{eq:normalized_signed_Laplacian}. Building on this idea, we propose below two specific graph network designs for signed graphs: Spectral-SGCN-I and Spectral-SGCN-II. The former behaves like a low-pass filtering and can be viewed as a signed graph counterpart of the vanilla GCN~\citep{KipWel17}. The latter is able to retain high-frequency information and can be viewed as a signed graph counterpart of FAGCN~\citep{BoXiaChu21}.
% Based on the frequency analysis of signed graph, one can extend the existing spectral GNNs for unsigned graphs to signed graphs. In this paper, we propose two extensions, namely Spectral-SGCN-I and Spectral-SGCN-II. 
%\chen{The parameter $\epsilon$ you mentioned seems to introduce some flexibility to the algorithm. This should be discussed.}

%----------------------
\subsection{Spectral-Signed-GCN-I}
Our first network design, Spectral-SGCN-I, is similar to the vanilla GCN~\citep{KipWel17}. It can be viewed as a low-pass feature aggregation on the underlying signed graph followed by feature transformation. At each layer, the features are first aggregated via low-pass filter
$\bP = \Tilde{\bD}^{-1/2} \Tilde{\bA} \Tilde{\bD}^{-1/2}$ with $\Tilde{\bA} = \bA + \bI$ and $\Tilde{\bD} = \Bar{\bD} + \bI$. It resembles \eqref{eq:renormalization} but uses the signed Laplacian. Note that, just like \eqref{eq:renormalization}, we adopt the renormalization trick to improve numerical stability~\citep{KipWel17}. 

In more details, the aggregated features in $\ell^{th}$ layer is $\Bar{\bH}^{(\ell)} = \bP \bH^{(\ell-1)}$. Let $\Bar{\bH}^{(\ell)} = [\Bar{\bh}_1^{(\ell)}, \Bar{\bh}_2^{(\ell)}, \ldots, \Bar{\bh}_N^{(\ell)}]^T$, then the aggregation can be written in the message passing form
\begin{align}\label{eq:aggr_lowpass}
    \Bar{\bh}^{(\ell)}_i =&  \frac{1}{\Bar{d_i} + 1} \bh_i^{(\ell - 1)} + \sum_{j\in \cN_i^+} \!\! \frac{1}{\sqrt{(\Bar{d_i}+1) (\Bar{d_j}+1)}} \bh_j^{(\ell - 1)} - \sum_{k\in \cN_i^-}\!\!\! \frac{1}{\sqrt{(\Bar{d_i}+1) (\Bar{d_k}+1)}} \bh_k^{(\ell - 1)}. \nonumber 
    % =& 
    % % ~ \mathrm{AGG}_{j \in \cN_i} \left( (\bP \bH^{(\ell - 1)})_j \right)
    % (\bP \bH^{(\ell - 1)})_i.
\end{align}
After aggregation, the features are transformed via a learnable parameter matrix along with non-linearity to give node representation output in $\ell^{th}$ layer as $\bH^{(\ell)} = \sigma \left( \Bar{\bH}^{(\ell)}~ \bTheta^{(\ell)} \right)$.

\begin{figure}[t]
\centering
\begin{subfigure}[t]{0.45\textwidth}
\centering	
\includegraphics[scale=0.8]{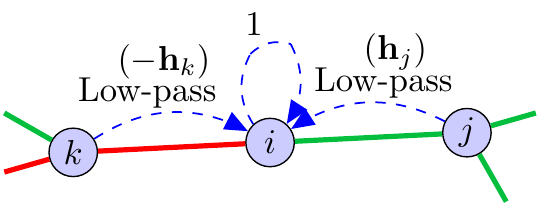}
\vspace{-0.2cm}
\caption{Spectral-SGCN-I.}
\label{fig:arch1}
\end{subfigure}
\begin{subfigure}[t]{0.45\textwidth}
\centering	
\includegraphics[scale=0.8]{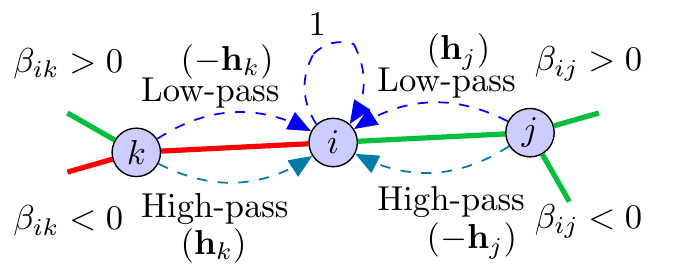}
\vspace{-0.2cm}
\caption{Spectral-SGCN-II.}
\label{fig:arch2}
\end{subfigure}
\vspace{-0.1cm}
\caption{Spectral-SGCN-I propagates only low-pass information from its neighbors. Spectral-SGCN-II propagates low-pass as well as high-pass information from its neighbors via attention coefficients.}
\label{fig:arch}
\end{figure}

The Spectral-SGCN-I aggregates only low frequency information via a low-pass aggregation filter as illustrated in Figure~\ref{fig:arch1}. As it has been shown in \citep{WuSouZha19} that removing nonlinearities and collapsing weight matrices between consecutive layers greatly simplifies the GCN complexity. Similarly, we propose spectral simplified signed graph convolution network (Spectral-S2GCN) such that the output of $\ell^{th}$ layer is $\bH^{(\ell)} = \bP^{\ell} \bX \bTheta$, with $\bTheta$ being the only learnable parameter matrix. 
% It has been shown in the unsigned GNN designs such as FAGCN~\citep{BoXiaChu21} that the high frequency information is also important and can improve performance. 

%*************************
%*************************

% \noindent \textbf{Spectral-SGCN-II:}\\
\subsection{Spectral-Signed-GCN-II}
As has been noted in FAGCN~\citep{BoXiaChu21}, besides low frequency components, it is beneficial to incorporate the high frequency components during feature aggregation as well. Based on this idea and the frequency interpretation on signed graphs, we extend FAGCN to signed graphs by considering low as well as high frequency information. To this end, we use low pass filter $\bP^{\mathrm{Low}} = \bI + \Bar{\bD}^{-1/2} \bA \Bar{\bD}^{-1/2} = 2\bI - \Bar{\bL}_{\mathrm{n}}$ and high pass filter $\bP^{\mathrm{High}} = \bI - \Bar{\bD}^{-1/2} \bA \Bar{\bD}^{-1/2} = \Bar{\bL}_{\mathrm{n}}$ along with attention to learn the proportion of low-frequency and high-frequency features to be propagated.

Let $\beta_{ij}^{\mathrm{Low}}$ be the coefficient of attention aggregation for low frequency features from node $j$ to node $i$. Similarly, let $\beta_{ij}^{\mathrm{High}}$ be the coefficient of attention aggregation for high frequency features from node $j$ to node $i$. Note $\beta_{ij}^{\mathrm{Low}}=\beta_{ij}^{\mathrm{High}}=0$ if node $i$ is not connected to node $j$. For target node $i$, define low-pass attention matrix as $\bB_{i}^{\mathrm{Low}} = \mathrm{diag}\{\beta_{i1}^{\mathrm{Low}},\ldots, \beta_{iN}^{\mathrm{Low}}\}$ and high-pass attention matrix as $\bB_{i}^{\mathrm{High}} = \mathrm{diag}\{\beta_{i1}^{\mathrm{High}},\ldots, \beta_{iN}^{\mathrm{High}}\}$. Let $\bH^{(\ell - 1)} = [\bh_1^{(\ell-1)}, \bh_2^{(\ell-1)},\ldots, \bh_N^{(\ell-1)}]^T$ be the node embeddings at layer $\ell -1$, then the $\ell^{th}$ GNN layer reads (assuming self-loops) 
\begin{align*}
    \bh^{(\ell)}_i &= \frac{1}{2} \left( \bP^{\mathrm{Low}} \bB_{i}^{\mathrm{Low}} \bH^{(\ell - 1)}\right)_i \!+\! \frac{1}{2} \left(\bP^{\mathrm{High}} \bB_{i}^{\mathrm{High}} \bH^{(\ell - 1)}\right)_i  \\
    &= (\beta_{ii}^{\mathrm{Low}} + \beta_{ii}^{\mathrm{High}}) \bh_i^{(\ell - 1)}\! + \!\!\!\! \sum_{j \in \cN_i^+} \!\! \frac{(\beta_{ij}^{\mathrm{Low}} - \beta_{ij}^{\mathrm{High}})} {\sqrt{\Bar{d}_i\Bar{d}_j }} \bh_j^{(\ell - 1)} - \sum_{k \in \cN_i^-}  \frac{(\beta_{ik}^{\mathrm{Low}} - \beta_{ik}^{\mathrm{High}})} {\sqrt{\Bar{d}_i\Bar{d}_k }} \bh_k^{(\ell - 1)}. 
\end{align*}

The coefficient $\beta_{ij}^{\mathrm{Low}} + \beta_{ij}^{\mathrm{High}}$ acts as a scaling factor and can be set to be $1$ for simplicity. Now denote $\beta_{ij}^{\mathrm{Low}} - \beta_{ij}^{\mathrm{High}} = \beta_{ij}$, then the above becomes
\begin{align*}
    \bh^{(\ell)}_i &=   \bh_i^{(\ell - 1)} \!+\!\! \sum_{j \in \cN_i^+}  \frac{\beta_{ij}} {\sqrt{\Bar{d}_i\Bar{d}_j }} \bh_j^{(\ell - 1)}  \!-\!\! \sum_{k \in \cN_i^-}  \frac{\beta_{ik}} {\sqrt{\Bar{d}_i\Bar{d}_k }} \bh_k^{(\ell - 1)}.
\end{align*}
%One can also consider taking $\beta_{ij}^{\mathrm{Low}}$ and $\beta_{ij}^{\mathrm{High}}$ separately.

When the attention coefficients are constant and equal to 1, the above reduces to Spectral-SGCN-I. In Spectral-SGCN-II, the attention coefficients are learned as $\beta_{ij} = \mathrm{tanh}\left(\ba^T [\bh_i,\bh_j] \right)$ taking values in range $[-1,1]$, where $\ba$ is learnable linear parameter. When $\beta_{ij} = \beta_{ij}^{\mathrm{Low}} - \beta_{ij}^{\mathrm{High}} > 0 $, the low-frequency information is propagated from node $j$ to node $i$ and when $\beta_{ij} < 0$, the high frequency information is propagated, as illustrated in Figure~\ref{fig:arch2}. Before passing the given input features $\bX$ to the first layer, they are first transformed to get $\bh_i^{(0)} = \sigma \left(\bTheta_1 \bx_i \right)$ and after $L$ number of stacked layer, we get the final output embeddings as $\bh_i = \bTheta_2 \bh_i^{(L)}$. 
% Here, $\sigma$ represents non-linearity and $\bTheta_1$ and $\bTheta_2$ are learnable linear transformations.

% In the undirected and unsigned case, when the adjacency matrix $\bA$ is symmetric with all non-negative entries, the (unnormalized) graph Laplacian can be defined by $\bL = \bD - \bA$, where $\bD = \mathrm{diag}\{d_1,d_2,\ldots,d_N\}$ is the diagonal degree matrix with $d_i = \sum_{j} A(i,j) \geq 0$. The Laplacian matrix $\bL$ is symmetric positive semidefinite and its eigendecomposition is used for spectral (frequency) analysis of the underlying unsigned and undirected graph. However when the graph is signed, $\bL$ no longer remains positive semidefinite and \singh{poses difficulties in the frequency analysis due to computational irregularities and ambiguous frequency ordering (see Section?)}. In order to handle signed graphs, the signed Laplacian can be used. \singh{Moreover, when the graph is directed ($\bA$ is asymmetric),  $\bL$ typically yields complex eigenvalues hindering straightforward extension of classical graph signal processing methods to directed graphs. For directed unsigned graphs, one can turn to the magnetic Laplacian. }
%%%%%%%%%%%%%%%%%%%%%%%%%%%%%%%%%%%%%%%%%%%%
\section{Directed Signed Graphs} \label{sec:directed_signed}

The methods proposed in Section~\ref{sec:proposed} are limited to undirected signed graphs. For unsigned graphs, the magnetic Laplacian ~\citep{FanAlaSuy17,FanAlaFer18,FurShiAki19} has been utilized to encode the edge directionality information. Recently, \citet{ZhaHeBru21} used magnetic Laplacian for designing GNNs for directed (unsigned) graphs. In its original form, the magnetic Laplacian is defined as 
\begin{equation}\label{eq:unnormalized_sL}
    \bL^q = \bD - \bA^q = \bD - \bA_s \odot \bPhi^q,
\end{equation}
where $\bA_s$ is symmetric adjacency matrix with entries $\bA_s(i,j) = \frac{1}{2}(\bA(i,j) + \bA(j,i))$, $\bD = \mathrm{diag}\{d_1,d_2,\ldots,d_N\}$ with $d_i = \sum_{j=1}^{N}\bA_s(i,j)$. Moreover, $\bPhi^q$ is a Hermitian matrix with elements 
\begin{equation}
    \bPhi^q(i,j) = e^{\iota 2\pi q (\bA(i,j) - \bA(j,i))},
\end{equation}
where $\iota$ is an indeterminate satisfying $\iota^2 = -1$ and $q \in [0,0.50)$ is the phase parameter. The normalized magnetic Laplacian is $\bL_{\mathrm{n}}^q = \bD^{-1/2} \bL^q \bD^{-1/2} $.

It can be shown that $\bL^q$ as well as $\bL_{\mathrm{n}}^q$ are positive semidefinite for the unsigned case and the eigenvalues of $\bL_{\mathrm{n}}^q$ lie in the interval $[0,2]$. However, when the underlying graph is signed, the degree matrix can have zero diagonal entries and the normalized magnetic Laplacian is not well defined. Moreover, $\bL^q$ becomes indefinite matrix for signed graphs. To handle these issues in directed signed graphs, we introduce \textit{signed magnetic Laplacian}. 
% Magnetic Laplacian for unsigned graphs 
% \begin{itemize}
%     \item Magnetic Eigenmaps for community detection in directed networks~\citep{FanAlaSuy17}
%     \item Magnetic Eigenmaps for the visualization of directed networks~\citep{FanAlaFer18}
%     \item Graph signal processing for directed graphs based on Hermitian Laplacian~\citep{FurShiAki19}
% \end{itemize}
%========================================
\subsection{Signed Magnetic Laplacian} \label{subsec:signed_mag_Laplacian}
\begin{figure}[ht]
\centering
\begin{subfigure}[t]{0.48\textwidth}
\centering	
\includegraphics[scale=0.76]{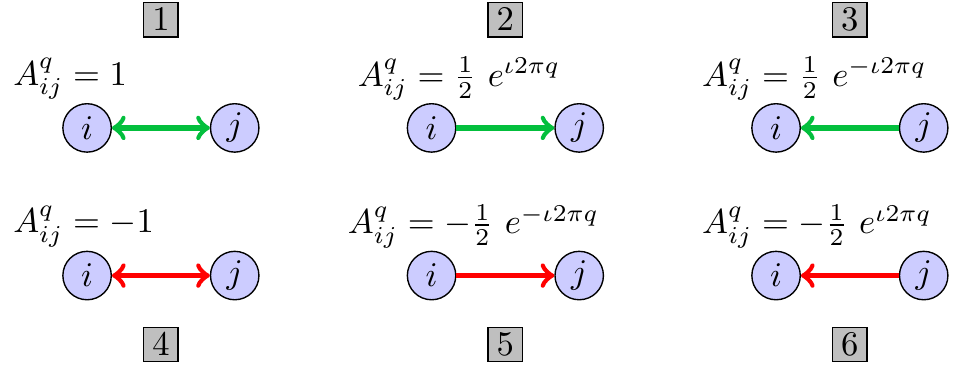}
% \caption{Different signed directed cases and corresponding $\bA^q$ values.}
\end{subfigure}
\hspace{0.2cm}
\begin{subfigure}[t]{0.48\textwidth}
\centering
\includegraphics[scale=0.76]{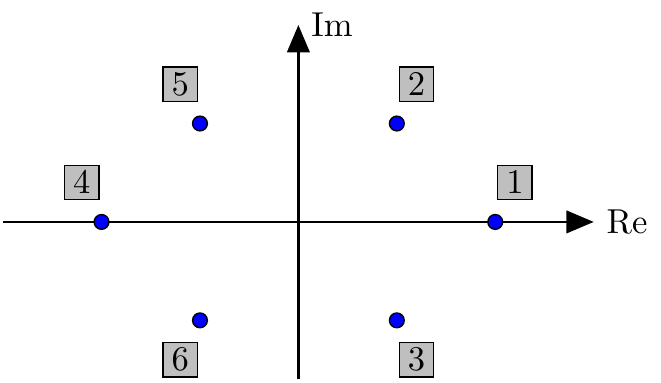}
% \caption{Singed directional phase locations in complex plane with $q=0.125$.}
\end{subfigure}
\caption{Illustration of different directed signed scenarios and singed directional phase locations on complex plane with $q=0.125$.}
\label{fig:direction1}
\end{figure}

We define signed magnetic Laplacian as 
\begin{equation}\label{eq:unnormalized_smL}
    \sL := \bar{\bD} - \bA^q =  \bar{\bD} - \bA_s \odot \bPhi^q,
\end{equation}
where $\bA^q = \bA_s \odot \bPhi^q$ contains the directional signed information via $\bPhi^q$ and the degree matrix is considered to be $\bar{\bD} = \mathrm{diag}\{\bar{d}_1,\bar{d}_2,\ldots,\bar{d}_N\}$ with $\bar{d}_i = \sum_{j=1}^{N}|\bA_s(i,j)|$ representing the connection strength of node $i$. Note that the phase parameter $q \in [0,0.25)$ for this signed directed settings (see Figure~\ref{fig:direction1} for illustration of different scenarios). The normalized signed magnetic Laplacian is $\nsL := \bar{\bD}^{-1/2} \sL \bar{\bD}^{-1/2}$. The eigendecomposition of $\nsL$ can be used for spectral analysis of directed signed graphs and directed signed convolution operations can be defined as in Section~\ref{subsec:traditional_GNNs}.

%*************************
%*************************

%---------------------------------------
\begin{prop} \label{prop:semidef}
The signed magnetic Laplacian $\sL$ and normalized signed magnetic Laplacian $\nsL$ are positive semidefinite. 
\end{prop}
%--------------------------------------
\begin{prop} \label{prop:range}
The eigenvalues of the normalized signed magnetic Laplacian $\nsL$ lie in $[0,2]$.
\end{prop}
%---------------------------------------
% %%%%%%%%%%%%%%%%%%%%%%%%%%%%%%%%%
% \subsection{Spectral Analysis of Signals on Directed Signed Graphs}
% %++++++++++++++++++++++
% \subsubsection{Directed Signed Graph Fourier Transform} 
% \label{subsubsec:gft}
% \begin{equation}
%     \bL_q = \bU \bLambda \bU^*,
% \end{equation}
% Define GFT, then define TV and frequency ordering
% %++++++++++++++++++++
% \subsubsection{Spectral Convolution and Filtering} 
% \label{subsubsec:filtering}
% Let $\bx, \by \in \mathbb{R}^N$ be two signals defined on a directed signed graph, then the directed signed convolution is defined as 

% \begin{align}
%     \bx ~\star~ \by = \bU \left( \hat{\by} \odot \hat{\bx} \right)
% \end{align}

%%%%%%%%%%%%%%%%%%%%%%%%%%%%%%%%%
\subsection{Directed Signed Graph Convolution Network}
Based on signed magnetic Laplacian, similar to MagNet~\citep{ZhaHeBru21}, we propose Signed-MagNet. In its aggregation operation, Signed-MagNet aggregates features by performing low-pass filtering as
\begin{align}\label{eq:aggr_lowpass}
    \Bar{\bh}^{(\ell)}_i =&  \frac{1}{\Bar{d_i} + 1} \bh_i^{(\ell - 1)} + \sum_{j\in \cN_i}  \frac{\bA^q(i,j)}{\sqrt{(\Bar{d_i}+1) (\Bar{d_j}+1)}} \bh_j^{(\ell - 1)},
\end{align}
where $\cN_i$ is the set of all the nodes connected to/from node $i$. Note that the latent embeddings in signed-MagNet are complex and at the last layer, we concatenate the real and imaginary parts. After aggregation operation in each layer, the features are transformed via a learnable complex matrix $\bTheta$. 

% After aggregation, the features are transformed via a learnable parameter matrix along with non-linearity to give node representation output in $\ell^{th}$ layer
% \begin{equation}
%     \bH^{(\ell)} = \sigma \left( \Bar{\bH}^{(\ell)}~ \bTheta^{(\ell)} \right).
% \end{equation}

% %*************************
% \begin{figure}[ht]
% \centering
% \includegraphics[scale=0.76]{./figures/aggregation.pdf}
% \caption{Aggregation illustration.}
% \label{fig:aggregation}
% \end{figure}
% %*************************
%--------------------------------------
\begin{table*}[ht]
\scriptsize
\caption{Signed graph datasets.}
\vspace{-0.4cm}
\begin{center}
\begin{tabular}{|c|c|c|c|c|c|c|}
\hline
   & Wiki-Editor & Wiki-Elections & Wiki-RfA & Bitcoin-Alpha & Bitcoin-OTC & Slashdot  \\
\hline\hline
  \#Nodes & 21535 & 7194  &  11381 & 3783 & 5881 &  79120  \\
  \#Classes & 2 & 2 & 2 & - & - & - \\
%   \#Links & 694436 & 203326  &  346217 \\
  \#Positive Links & 269251 & 81862  & 139345  & 22650 & 32029 &  392179 \\ 
  \#Negative Links & 79004 &  22497 & 39433 & 1563 & 3563 &  123218 \\
   \%Positive Links & 77.31 \% & 78.44 \% & 77.94 \% & 93.54 \% & 89.99 \% & 76.09 \% \\
\hline
\end{tabular}
\end{center}
\label{table:datasets}
\end{table*}
%_-----------------------------------------
%%%%%%%%%%%%%%%%%%%%%%%%%%%%%%%%%%%%%%%%%%%%%%%
\section{Experiments}\label{sec:experiments}
We evaluate our proposed methods for node classification and link sign predictions tasks on signed networks. We used Deep Graph Library (DGL)~\citep{dgl} for implementation of our methods. \singh{We also utilized PyTorch Geometric Signed Directed~\citep{HeZhaHua22} for implementing existing signed GNN baselines for node classification tasks.} 
% All the experiments were run on Intel Core i9-9900 machine equipped with one NVIDIA GeForce RTX 2080 Ti GPU.
%===========================
\subsection{Datasets} \label{subsec:datasets}
We perform node classification task on three datasets: Wiki-Editor, Wiki-Election, and Wiki-RfA. Wiki-Editor is extracted from the UMDWikipedia dataset~\citep{KumSpeSub15}. 
% The dataset consists of vandals and benign editors of Wikipedia.
There is a positive edge between two users if their co-edits belong to the same categories and a negative edge represents the co-edits belonging to different categories. Each node is labeled as either benign or vandal. Wiki-RfA~\citep{WesPasLes14} and Wiki-Election~\citep{LesHutKle10} are datasets of editors of Wikipedia that request to become administrators. A request for adminship (RfA) is submitted, either by the candidate or by another community member and any Wikipedia member may give a supporting, neutral, or opposing vote. From these votes a signed network is built for each dataset, where a positive (resp. negative) edge indicates a supporting (resp. negative) vote by a user and the corresponding candidate. The label of each node in these networks is given by the output of the corresponding request: positive (resp. negative) if the editor is chosen (resp. rejected) to become an administrator. 
We use dataset extraction code provided by \citet{MerBosSto19}~\footnote{https://github.com/melopeo/GL}.
For link sign prediction, we use three additional datasets: Bitcoin-Alpha, Bitcoin-OTC, and Slashdot~\footnote{https://networks.skewed.de}. Bitcoin-Alpha and Bitcoin-OTC~\citep{KumSpeSub16,KumHooMak18} are two exchanges in Bitcoins, where nodes represent Bitcoin users and edges represent the level of trust/distrust they have in other users. Slashdot dataset~\citep{KunLomBau09} is a network of interactions among users on Slashdot. Nodes represent users and edges represent friends (positive) or foes (negative). The dataset statistics are summarized in Table~\ref{table:datasets}.

%%%%%%%%%%%%%%%%%%%%%%%%%%%%%%%%%%%%%
\subsection{Results}
\label{subsec:Results}
\begin{figure}[ht]
\centering
\begin{subfigure}[t]{0.45\textwidth}
\centering	
\includegraphics[scale=0.4]{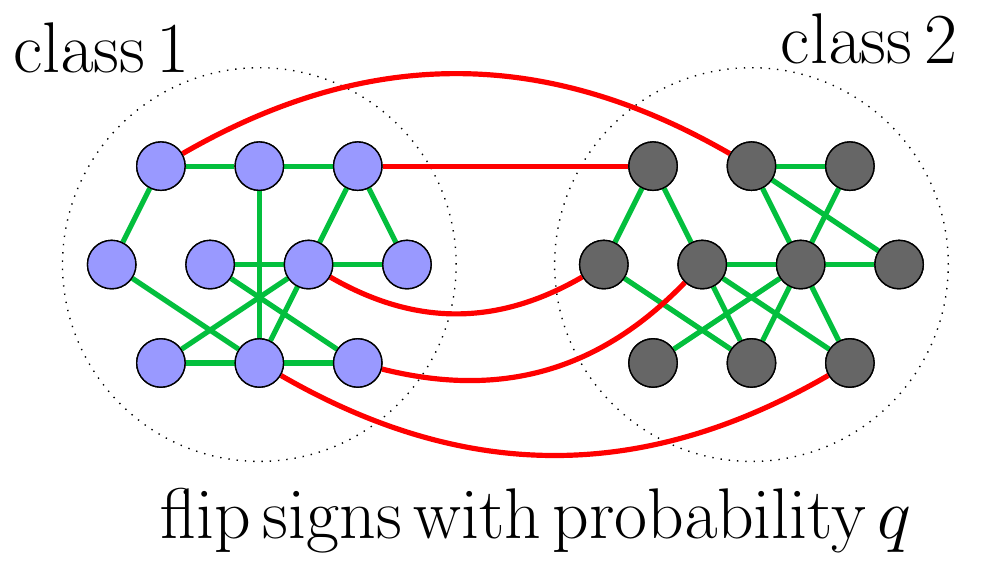}
\end{subfigure}
\begin{subfigure}[t]{0.45\textwidth}
\centering
\includegraphics[scale=0.21]{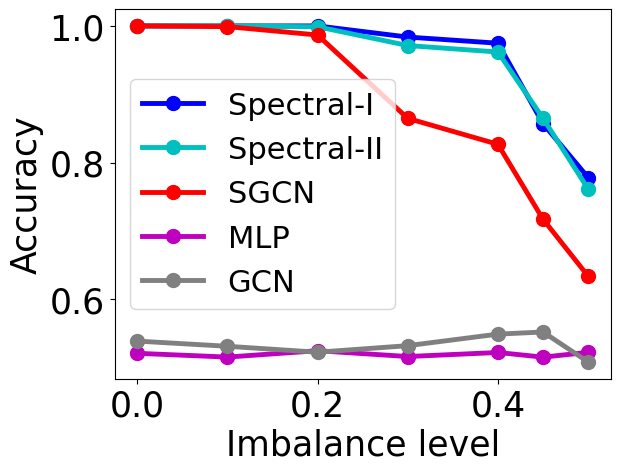}
\end{subfigure}
\vspace{-0.2cm}
\caption{Performance with varying levels of imbalance.}
\label{fig:imbalance}
\end{figure}

%-----------------------------------
% \subsubsection{Synthetic Data}
% \label{subsubsec:synthetic_results}
We first present experiments on synthetic data generated by signed stochastic block model~\citep{CucDavGli19, HeReiWan22} with different levels of imbalance. We simulate two clusters with intra-cluster edge probability of 0.02 having positive signs and inter-cluster probability of 0.01 having negative signs. Such a graph is a balanced graph. We then flip the inter-cluster as well as intra-cluster edge signs with different probabilities creating varying levels of imbalance. With only 2 node labels known per cluster for a total of 1000 nodes, the node classification performance is shown in Figure~\ref{fig:imbalance} (one hot vectors as input features and hidden dimension of 16). Clearly, our proposed methods outperform SGCN significantly.
%*************************
%==================================
%--------------------------------------
\begin{table*}[ht]
\setlength{\tabcolsep}{5pt}
\renewcommand{\arraystretch}{1.7}
\tiny
\caption{\singh{Test accuracies averaged over 20 different seeds for data shuffles and initializations.}}
\vspace{-0.4cm}
\begin{center}
\begin{tabular}{|c|c|c|c|c|c|c|c|c|c|}
\hline
% \multirow{2}{*}{Dataset} &
Dataset &  \multicolumn{3}{c|}{WikiEditor} & \multicolumn{3}{c|}{WikiElection} & \multicolumn{3}{c|}{WikiRfA}                   \\
\hline
  Known Labels & 1 \% & 2 \% & 5 \%  & 1 \% & 2 \% & 5 \%  & 1 \% & 2 \% & 5 \%  \\
\hline\hline
MLP & 72.3 $\pm$ 1.7  & 73.6 $\pm$ 1.1 & 74.8 $\pm$ 0.7   & 90.3 $\pm$ 0.9 & 93.7 $\pm$ 0.8  & 95.5 $\pm$ 0.6 & 91.5 $\pm$ 0.8 & 92.4 $\pm$ 1.2  & 93.8 $\pm$ 0.5   \\
   \hline
  \singh{GCN} & 73.0 $\pm$ 0.8 & 73.3 $\pm$ 0.9 & 74.5 $\pm$ 1.0 &  88.9 $\pm$ 3.4 & 89.8 $\pm$ 3.7 & 89.9 $\pm$ 3.8 & 88.1 $\pm$ 2.2  & 89.4 $\pm$  2.2  & 90.0 $\pm$ 2.8  \\
  \hline
 SAGE & 70.3  $\pm$ 0.8 & 72.8  $\pm$ 0.9 &  76.1 $\pm$ 0.5   & 87.2  $\pm$ 1.2 &  89.7  $\pm$ 1.2 & 93.3 $\pm$ 0.7 & 87.9 $\pm$ 0.9 & 89.9  $\pm$ 0.8 & 93.3 $\pm$ 0.6  \\
  \hline
  FAGCN & 75.4 $\pm$ 1.1 & 77.2 $\pm$ 0.9 & 78.8 $\pm$ 0.5   & 92.1 $\pm$ 1.4 & 93.3 $\pm$ 0.9 & 95.4 $\pm$ 0.7 & 91.5 $\pm$ 1.0 & 92.4 $\pm$ 0.8 & 94.1 $\pm$ 0.4  \\
  \hline
  SGCN & 79.4 $\pm$ 0.7  & 80.3 $\pm$ 0.9 & 82.1 $\pm$ 0.4   & 87.5 $\pm$ 1.2 & 89.9 $\pm$ 0.7 & 93.8 $\pm$ 0.4 & 87.9 $\pm$ 1.2 & 89.5 $\pm$ 1.7 & 91.3 $\pm$ 1.5  \\
   \hline
   \singh{SNEA} & 79.5 $\pm$ 1.1  & 79.9 $\pm$ 1.0 & 81.1 $\pm$ 1.3   & 93.1 $\pm$ 1.5 & 94.1 $\pm$ 1.7 & 95.2 $\pm$ 1.7 & 93.0 $\pm$ 1.0 & 94.0 $\pm$ 1.1 & 94.4 $\pm$ 1.2  \\
   \hline
   \singh{SDGNN} & 80.5 $\pm$ 1.3  & 81.1 $\pm$ 0.9 & 82.2 $\pm$ 0.7   & 95.8 $\pm$ 1.0 & 96.1 $\pm$ 0.6 & 96.7 $\pm$ 0.8 & 94.8 $\pm$ 1.1 & 95.0 $\pm$ 0.7 & 95.3 $\pm$ 0.4  \\
   \hline
   Spectral-SGCN-I & \textbf{81.1 $\pm$ 0.8} & 81.3 $\pm$ 0.8 & 82.8 $\pm$ 0.4 &  95.7 $\pm$ 1.0 & 97.1 $\pm$ 0.5 & 97.6 $\pm$ 0.4 & 94.4 $\pm$ 0.7 & 95.1 $\pm$ 0.5 & 95.7 $\pm$ 0.6  \\
  \hline
  Spectral-SGCN-II & 80.0 $\pm$ 2.2  & 80.9 $\pm$ 1.5 & 82.3 $\pm$ 1.4 &  96.7 $\pm$ 1.0 & 97.4 $\pm$ 0.3 & \textbf{98.0 $\pm$ 0.3} & 94.9 $\pm$ 0.4 & 95.6 $\pm$ 0.4 & 96.0 $\pm$ 0.2  \\ \hline
   Spectral-S2GC & 81.1 $\pm$ 0.8  & 81.8 $\pm$ 0.7 & 82.7 $\pm$ 0.7  & 96.6 $\pm$ 0.9 & 97.3 $\pm$ 0.5 &  97.8 $\pm$ 0.2 &  94.3 $\pm$ 0.7 &  95.2 $\pm$ 0.4 &  95.9 $\pm$ 0.2  \\
  \hline
   Signed-MagNet & 81.0 $\pm$ 1.1  & \textbf{82.0 $\pm$ 0.8} & \textbf{83.0 $\pm$ 0.4} & \textbf{97.2  $\pm$ 0.6}  & \textbf{97.6 $\pm$ 0.3} &  98.0 $\pm$ 0.2  & \textbf{96.0  $\pm$ 0.4}  & \textbf{96.2 $\pm$ 0.3} &  \textbf{96.4 $\pm$ 0.2}  \\
\hline
\end{tabular}
\end{center}
\label{table:node_classification1}
\end{table*}
%----------------------------------------
%-----------------------------------
\subsubsection{Node Classification}
\label{subsubsec:Node_class_results}
Next, we perform node classification task in a semi-supervised setting, i.e., we have access to the test data, but not the test labels, during training. For all the three datasets, we use three different ratios for training (known labels): $1\%,~2\%,~5\%$ of the total nodes. Out of the remaining nodes, we use $90\%$ for testing, and use rest of the nodes for validation. Since the features are not given for the nodes, we use truncated SVD of the symmetric adjacency matrix with dimension of 64 as input features. % as has been done in signed GNN models including SGCN~\citep{DerMaTan18} and SNEA~\citep{LiTiaZha20}. 
For comparison, we use traditional unsigned GNN designs namely GCN~\citep{KipWel17}, GraphSAGE~\citep{HamYinLes17}, and FAGCN~\citep{BoXiaChu21}. For these methods, we do not consider the sign of the links, since the signed edge information is not applicable for these methods. \singh{We also use the state-of-the-art GNN designs based on balance theory including SGCN~\citep{DerMaTan18}, SNEA~\citep{LiTiaZha20}, and SDGNN~\citep{HuaSheHou21} for comparison.} For fair comparison, we use two layer networks with hidden dimension of 64 for all the GNN-based methods. Binary cross-entropy loss based on the known labels is used as a loss function. We use ReLU as the non-linearity function in between the layers. Adam is used as the optimizer along with $\ell_2$-regularization to avoid overfitting. We tune the learning rate and weight decay ($\ell_2$-regularization) hyperparameters over validation data using a grid search. For Signed-MagNet implementation, we fix $q = 0.125$ for all the experiments. Further details on implementation and hyperparameter tuning are provided in Appendix.

The classification results are summarized in Table~\ref{table:node_classification1}. The experiments were run for 300 epochs and the results are averaged over 20 different random splits of training and test data. The average of best accuracies along with the standard deviation over 20 runs is reported. The best performing model for each dataset is in bold. We observe that our proposed signed GNNs consistently outperform the other methods in all the three datasets. For the two datasets Wiki-Election and Wiki-RfA, even traditional methods without signed information outperform SGCN. 

%=========================================
\subsubsection{Link Sign Prediction}
\label{subsubsec:link_pred}

Finally, we evaluate our methods for the task of link sign prediction that aims to predict the missing sign of a given edge. There exist three type of links in a signed graph: positive link, negative link, and no link. Denote this as a set $\cS \in \{+,-,?\}$, with $``?"$ representing no link. Specifically, the training data contains a set of nodes $\cV_t$ and a set of link triplets $\cT$ consisting of triplets of the form $(u,v,s_{uv})$ with $u,v \in \cV$ being node pairs and $s_{uv} \in \cS$ denoting type of link between $u$ and $v$. The final embeddings (obtained from the GNN model) of the two nodes $u$ and $v$ are concatenated together $[\bh_u, \bh_v]$ as the set of features for the edge and then fed to a three-class MLP classifier. The models are trained using the labeled edges from the training data. Let one-hot encoded vector of link type $s_{uv}$ be $\bs_{uv} \in \{0,1\}^{|\cS|}$. We use multi-class (three) cross entropy loss: 
\vspace{-0.3cm}
\begin{equation*}
    \cL(\bTheta, \bW) = - \frac{1}{|\cT|} \sum_{(u,v,s_{uv}) \in \cT} \sum_{c=1}^{|\cS|} \bs_{uv}(c) ~\mathrm{log} (\hat{\bs}_{uv} (c)),
\end{equation*}
where $\hat{\bs}_{uv} (c)$ is the predicted probability for class $c$ via $\mathrm{softmax}$ function. In the above loss function, $\bTheta$ denotes the set of GNN parameters and $\bW$ denotes the parameters of MLP classifier.  We use $80\%$ of the links for training and rest $20\%$ for testing. We use twice the number of training links as ``no links" obtained using negative sampling. 

% We use SiNE~\citep{WanTanAgg17}~\footnote{https://suhangwang.ist.psu.edu/codes/SiNE.zip}, SLF~\citep{XuHuWu19}~\footnote{https://github.com/WHU-SNA/SLF}, SGCN~\citep{DerMaTan18}~\footnote{https://github.com/benedekrozemberczki/SGCN}, and SNEA~\citep{LiTiaZha20}~\footnote{https://github.com/liyu1990/snea} 

\singh{We use SiNE~\citep{WanTanAgg17}, SLF~\citep{XuHuWu19}, SGCN~\citep{DerMaTan18}, SNEA~\citep{LiTiaZha20}, and SDGNN~\citep{HuaSheHou21} as baselines for comparison on link sign prediction tasks.} We use a two layer GNN model along with a single hidden layer MLP classifier.
% and use Adam optimizer along with $\ell_2$-regularization.
Truncated SVD of the symmetric adjacency matrix with dimension of 30 is used as input features. \singh{We utilize Marco-F1 and Micro-F1 scores for evaluation, since the positive and negative links are unbalanced. The comparison results in terms of F1 scores are listed in Table~\ref{table:link_prediction}. The results reported in the table are average F1 scores over 10 different runs.} We observe that our methods outperform the baselines for all the dataset except for Slashdot and WikiRfA. Note that all the numbers for other algorithms in the table are obtained by running the official codes provided by the respective authors. These could be slightly different from the numbers reported in those papers that may use filtered/truncated versions of datasets.

% We utilize Area Under the receiver operating characteristic Curve (AUC) for evaluation, since the positive and negative links are unbalanced. 
%For all the GNN based models, we use two layers with hidden dimensions of 64.

% \textbf{Issues When we Drop Direction Information:} If we drop the direction information, the magnetic Laplacian reduces to signed Laplacian. In such case, the clustering method described above has no way of discriminating the nodes within a cluster. 

% \multicolumn{2}{c}{Multi-column}

%-----------------------------------------
\begin{table*}[t]
\setlength{\tabcolsep}{6pt}
\renewcommand{\arraystretch}{1.4}
\tiny
\caption{\singh{Link prediction results with Macro-F1 and Micro-F1 scores over 10 different runs.}}
\vspace{-0.4cm}
\begin{center}
\begin{tabular}{|c|c|c|c|c|c|c|c|c|c|c|c|c|}
\hline
Method & \multicolumn{2}{|c|}{Bitcoin-Alpha} & \multicolumn{2}{|c|}{Bitcoin-OTC} & \multicolumn{2}{|c|}{Slashdot}  & \multicolumn{2}{|c|}{WikiElection}  & \multicolumn{2}{|c|}{WikiEditor}  & \multicolumn{2}{|c|}{WikiRfA}  \\  
\hline 
SiNE & 0.6790 & 0.9440 & 0.6832 & 0.9328 & 0.7238 & 0.8192 & 0.6940 & 0.7832 & 0.7858 & 0.8244 & 0.7263 & 0.8160 \\
SLF & 0.7475 & 0.9453 & 0.7483 & 0.9466 & \textbf{0.7943} & \textbf{0.8564} & 0.7616 & 0.8487 & 0.7521 & 0.8442 & \textbf{0.7881} & \textbf{0.8640} \\
SGCN & 0.6648 & 0.9184 & 0.7420 & 0.9012 & 0.7422 & 0.8260 & 0.7363 & 0.8360 & 0.8225 & 0.8548 & 0.7510 & 0.8430 \\
SNEA & 0.6796 & 0.9210 & 0.7580 & 0.9038 & 0.7431 & 0.8308 & 0.7388 & 0.8372 & 0.8290 & 0.8672 & 0.7542 & 0.8497 \\
\singh{SDGNN} & 0.7412 & 0.9480 & 0.8020 & 0.9346 & \underline{0.7820} & \underline{0.8504} & 0.7550 & 0.8508 & 0.8529 & 0.8810 & \underline{0.7832} & \underline{0.8598} \\
Spectral-SGCN-I & 0.7518 & \underline{0.9547} & \underline{0.8371} & 0.9437 & 0.7705 & 0.8410 & 0.7466 & 0.8390 & 0.8560 & 0.9128 & 0.7740 & 0.8410 \\
Spectral-SGCN-II & \underline{0.7630} & \textbf{0.9562} & 0.8356 & \underline{0.9538} & 0.7724 & 0.8438 & \underline{0.7645} & \underline{0.8512} & \textbf{0.8825} & \textbf{0.9237} & 0.7735 & 0.8476 \\
Spectral-S2GCN & 0.7030 & 0.9418 & 0.7820 & 0.9284 & 0.7352 & 0.8194 & 0.7182 & 0.8390 & 0.8448 & 0.9083 & 0.7221 & 0.8278 \\
Singned-Magnet & \textbf{0.7880} & 0.9432 & \textbf{0.8548} & \textbf{0.9580} & 0.7781 & 0.8463 & \textbf{0.7818} & \textbf{0.8649} & \underline{0.8652} & \underline{0.9184} & 0.7692 & 0.8543 \\
\hline
\end{tabular}
\end{center}
\label{table:link_prediction}
\end{table*}
\section{Conclusion}\label{sec:Conclusion}
In this paper, we proposed a new framework for GNN design for signed graphs based on spectral domain analysis over signed graphs, as opposed to existing balance theory based GNN methods. We also introduced signed magnetic Laplacian for handling directed signed graphs. We evaluated our methods for node classification as well as link sign prediction tasks on signed graphs and achieved state-of-the-art performance. 

\newpage
\bibliographystyle{nips22}
\bibliography{neurips_2022.bib}
\newpage 
\appendix

\section{Proof of Proposition 1}
%---------------------------------------
% \begin{prop} 
% The signed magnetic Laplacian $\sL$ and normalized signed magnetic Laplacian $\nsL$ are positive semidefinite. 
% \end{prop}
\begin{proof}
Let $\bX^{\dagger}$ be the conjugate transpose of $\bX$ and let $\bx \in \mathbb{C}^N$. It is easy to see that $\sL$ is a Hermitian matrix and therefore, the imaginary part $Im(\bx^{\dagger} \sL \bx) = 0$. Denoting $\bPhi^q(m,n) = e^{\iota \bTheta^q(m,n)}$, where $\bTheta^q(m,n) = 2\pi q (\bA(m,n) - \bA(n,m))$. The real part
\begin{align*}
    Re(\bx^{\dagger} \sL \bx) &=  \sum_{n=1}^N  \bar{\bD}(n,n) \bx(n) \bx^*(n)  -  \sum_{m,n=1}^N \bA_s(m,n) \bx(m) \bx^*(n) \cos(\bTheta^q(m,n)) \\
    &=   \sum_{m,n=1}^N  |\bA_s(m,n)| |\bx(m)|^2 -   \sum_{m,n=1}^N \bA_s(m,n) \bx(m) \bx^*(n) \cos(\bTheta^q(m,n)) \\
    % &= \frac{1}{2} \sum_{m,n=1}^N  |\bA_s(m,n)| |\bx(m)|^2 + \frac{1}{2}\sum_{n,m=1}^N  |\bA_s(n,m)| |\bx(n)|^2 -   \sum_{m,n=1}^N \bA_s(m,n) \bx(m) \bx^*(n) \cos(\bTheta^q(m,n)) \\
    &\geq  \frac{1}{2} \sum_{m,n=1}^N  |\bA_s(m,n)| |\bx(m)|^2 + \frac{1}{2} \sum_{n,m=1}^N  |\bA_s(n,m)| |\bx(n)|^2 \\
    & \quad \quad -   \sum_{m,n=1}^N |\bA_s(m,n) \bx(m) \bx^*(n) \cos(\bTheta^q(m,n))| \\
    &= \frac{1}{2} \sum_{m,n=1}^N  |\bA_s(m,n)| \left( |\bx(m)|^2 +  |\bx(n)|^2 -  2 |\bx(m) \bx^*(n) \cos(\bTheta^q(m,n))| \right) \\
    &\geq \frac{1}{2} \sum_{m,n=1}^N  |\bA_s(m,n)| \left( |\bx(m)|^2 +  |\bx(n)|^2 -  2 |\bx(m)| |\bx(n)| \right) \\
    &= \frac{1}{2} \sum_{m,n=1}^N  |\bA_s(m,n)| \left( |\bx(m)| -   |\bx(n)| \right)^2 \\
    &\geq 0.
\end{align*}
Letting $\by =\bar{\bD}^{-1/2} \bx$ and by definition of $\nsL$, we have
\begin{align*}
    \bx^{\dagger} \nsL \bx = \bx^{\dagger}  \bar{\bD}^{-1/2} \sL \bar{\bD}^{1/2}  \bx =  \by^{\dagger} \sL   \by &\geq 0.
\end{align*}

\end{proof}

\section{Proof of Proposition 2}
%--------------------------------------
% \begin{prop} 
% The eigenvalues of the normalized signed magnetic Laplacian $\nsL$ lie in the interval $[0,2]$.
% \end{prop}
\begin{proof}
Since $\nsL$ is positive semidefinite due to Proposition 1, we just show that the largest eigenvalue $\lambda_{N} \leq 2$. We know that the eigenvalue with largest absolute value is
\begin{align*}
    \lambda_N = \maxwrt[\bx \neq 0]{~\frac{\bx^{\dagger} \nsL \bx}{\bx^{\dagger} \bx}}.
\end{align*}
Letting $\by =\bar{\bD}^{-1/2} \bx$, we have
\begin{align*}
    \lambda_N = \maxwrt[\bx \neq 0]{~\frac{\bx^{\dagger} \bar{\bD}^{-1/2} \sL \bar{\bD}^{1/2}  \bx}{\bx^{\dagger} \bx}} = \maxwrt[\by \neq 0]{~\frac{\by^{\dagger} \sL \by}{\by^{\dagger} \bar{\bD} \by}}.
\end{align*}
Since the numerator in the above 
\begin{align*}
    \by^{\dagger} \sL \by &=  \sum_{m,n=1}^N  |\bA_s(m,n)| |\by(m)|^2 -   \sum_{m,n=1}^N \bA_s(m,n) \by(m) \by^*(n) \cos(\bTheta^q(m,n)) \\
    &= \frac{1}{2} \sum_{m,n=1}^N  |\bA_s(m,n)| |\by(m)|^2 + \frac{1}{2} \sum_{m,n=1}^N  |\bA_s(m,n)| |\by(n)|^2 \\
    & \quad \quad -   \sum_{m,n=1}^N \bA_s(m,n) \by(m) \by^*(n) \cos(\bTheta^q(m,n)) \\
    &\leq \frac{1}{2} \sum_{m,n=1}^N  |\bA_s(m,n)| \left (|\by(m)| +  |\by(n)| \right)^2 \\
    &\leq \sum_{m,n=1}^N  |\bA_s(m,n)| \left (|\by(m)|^2 +  |\by(n)|^2 \right)\\
    &= 2 \sum_{m,n=1}^N  |\bA_s(m,n)|  ~|\by(m)|^2 \\
    &= 2  \sum_{m=1}^N \left( \sum_{n=1}^N  |\bA_s(m,n)|\right)  ~|\by(m)|^2 \\
    &= 2 \sum_{m=1}^N \bar{\bD}(m,m) ~|\by(m)|^2 \\
    &= 2 \by^{\dagger} \bar{\bD} \by,
\end{align*}
and thus, $\lambda_N \leq 2$. 
\end{proof}
%---------------------------------------
%*************************
\begin{figure}[h]
\centering
\includegraphics[scale=0.2]{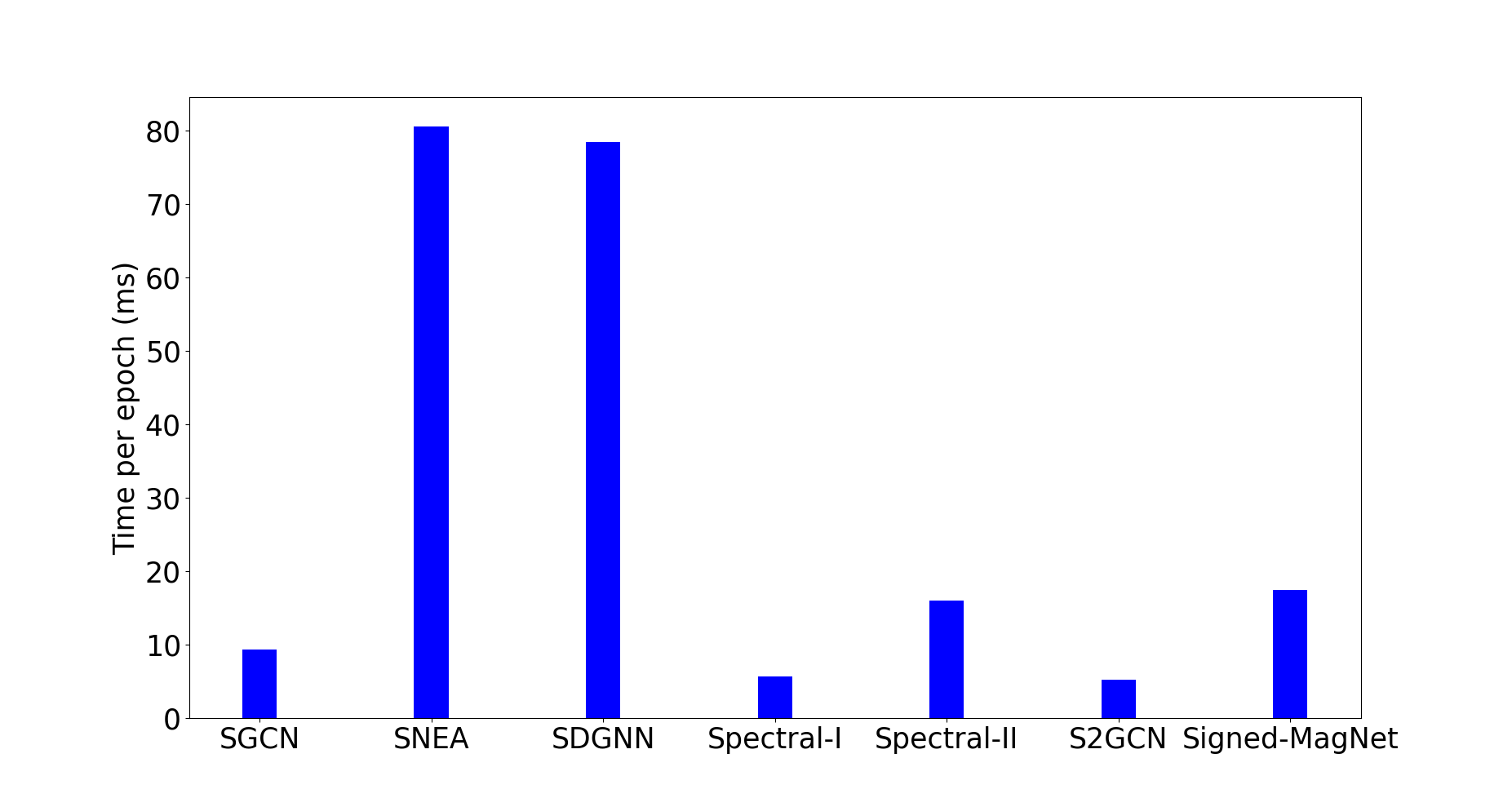}
\caption{\singh{Time taken per epoch for different methods (on WikiRfA dataset for node classification task with batch size = 64, number of GNN layers = 2, hidden dimension = 64).}}
\label{fig:time}
\end{figure}
%*************************
%=================================
\section{Further Implementation Details}
All the experiments were run on Intel Core i9-9900 machine equipped with NVIDIA GeForce RTX 2080 Ti GPU. 
We use two layer networks with hidden dimension of 64 for all the GNN-based methods (a standard practice in unsigned GNN literature). ReLU nonlinearity was used in all the experiments. For the implementation of Signed-Magnet, we used ReLU non-linearity for real and complex parts separately. The only hyperparameters to tune are learning rate and regularization (weight decay) coefficient. \singh{For all of our methods we use feature dropout with a rate of 0.5. For Spectral-SGCN-II, we use attention and feature dropout with dropout rate of 0.5}. We tune the learning rate with different values (on log scale) in the range $[1e^{-3}, 1e^{-1}]$ and regularization rate in the range $[1e^{-6}, 1e^{-3}]$. For node classification task, the hyperparameters were tuned based on the validation accuracy with $1\%$ known training labels for each dataset.

\singh{\textbf{Complexity Analysis:} Although the time complexity of the existing signed GNN designs (balance theory based) and our methods is $\mathcal{O}(|E|)$ for sparse graphs or $\mathcal{O}(|V|^2)$ in worst-case, the number of parameters per layer for these methods are different. For example, SDGNN utilizes four different encoders for capturing different directionality scenarios. In contrast, our signed spectral methods employ a single (transformation) encoder in each layer. We provide training times per epoch (averaged over 300 epochs) for different signed GNN methods in Figure~\ref{fig:time}.
}

%=============================
\section{Clustering of Directed Signed Graphs}

%*************************
\begin{figure*}[ht]
\centering
\begin{subfigure}[t]{0.32\textwidth}
\centering	
\includegraphics[scale=0.22]{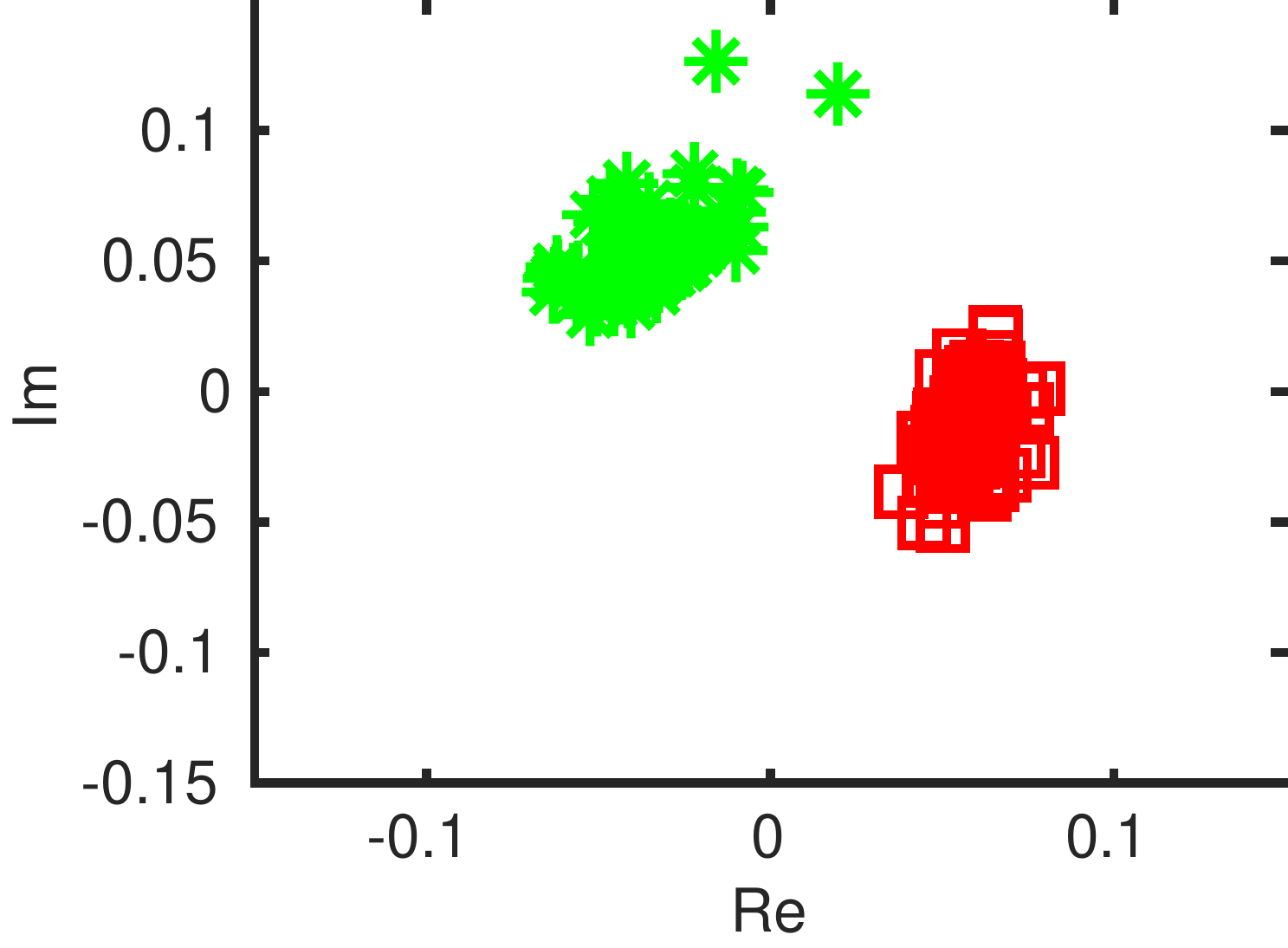}
\caption{imbalance prob. = 0.}
\end{subfigure}
\begin{subfigure}[t]{0.32\textwidth}
\centering
\includegraphics[scale=0.22]{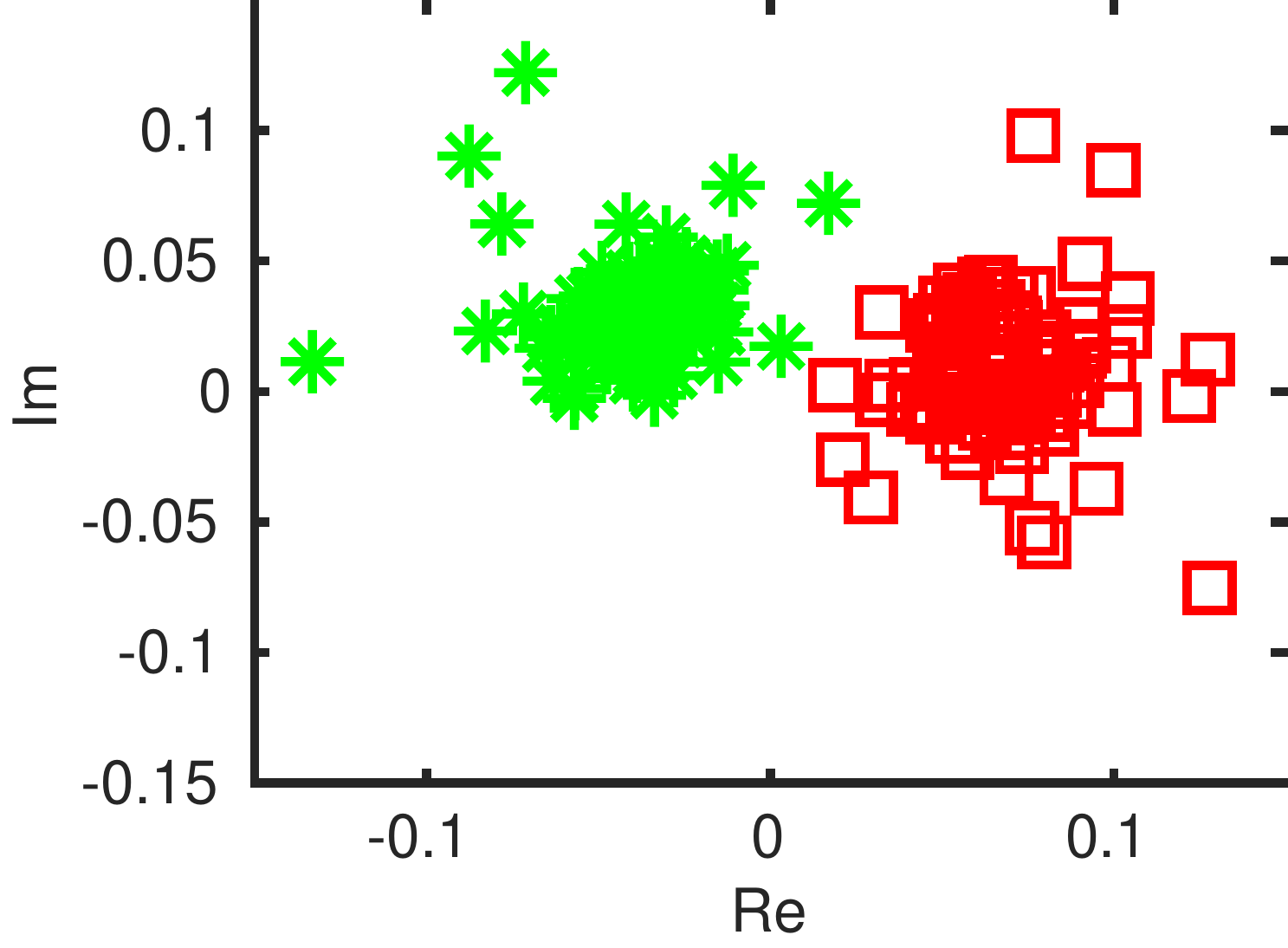}
\caption{imbalance prob. = 0.1.}
\end{subfigure}
\begin{subfigure}[t]{0.32\textwidth}
\centering
\includegraphics[scale=0.22]{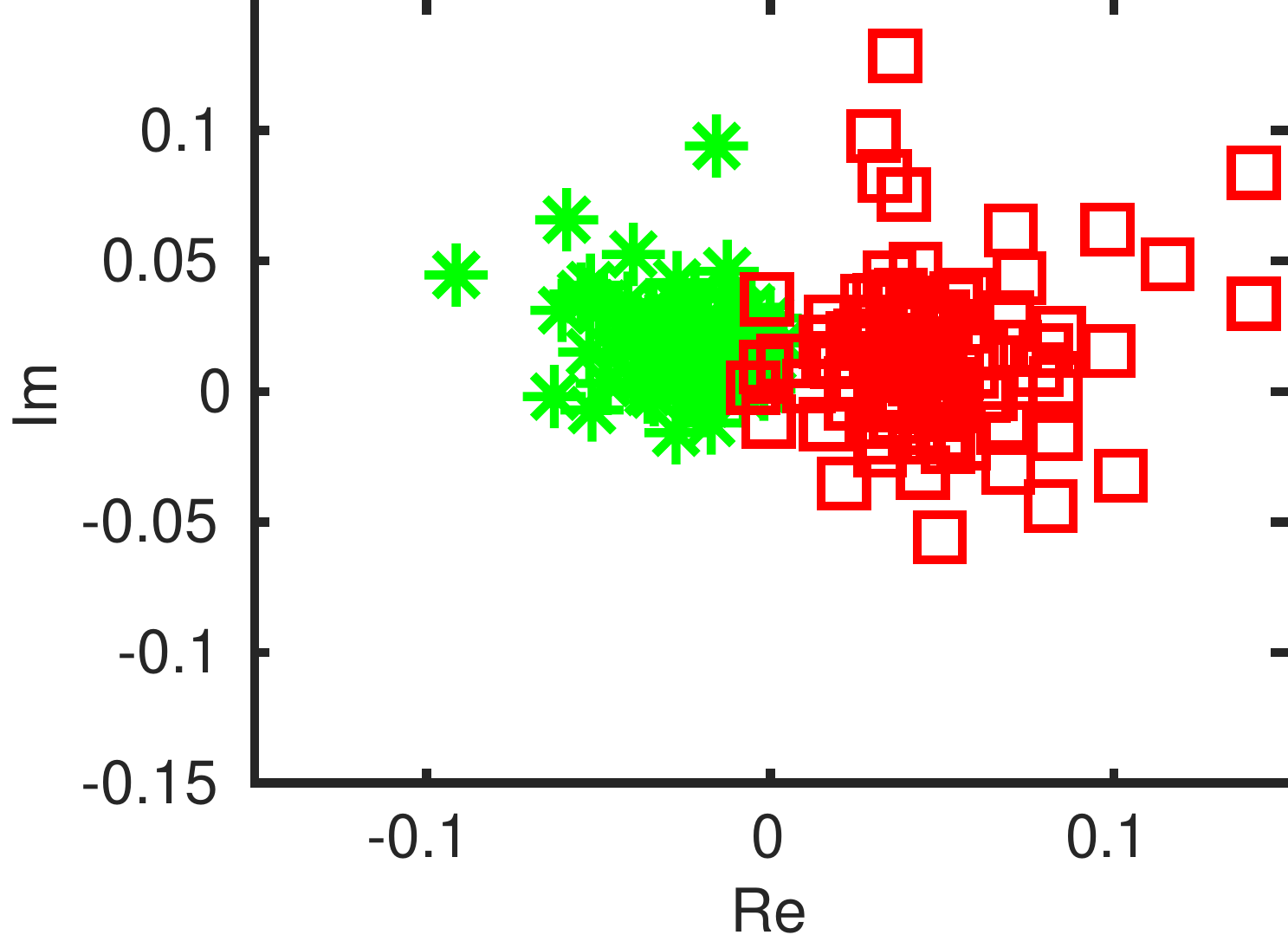}
\caption{imbalance prob. = 0.2.}
\end{subfigure}
\caption{Clustering results with 100 nodes per cluster for directed signed stochastic block model.}
\label{fig:clustering1}
\end{figure*}
%****************************

We present experiments on synthetic data generated by directed signed stochastic block model with different levels of imbalance. In particular, we simulate two clusters (classes) with directed intra-cluster edge probability of 0.05 having positive signs and directed inter-cluster edge probability of 0.05 having negative signs. Such a graph is a balanced graph. We then flip the inter-cluster as well as intra-cluster edge signs with different probabilities creating varying levels of imbalance.

The nodes can be clustered based on the first (complex) eigenvector of the signed magnetic Laplacian corresponding to the eigenvalue with smallest absolute value plotted on the complex plane. Figure~\ref{fig:clustering1} shows clustering results on directed signed graphs with varying levels of imbalance. 

%==============================
\section{Additional Experiments on Link Sign Prediction}

We present additional results on link sign prediction in terms of Area Under the receiver operating characteristic Curve (AUC) scores in Table~\ref{table:link_prediction_auc}.
%-----------------------------------------
\begin{table*}[t]
\setlength{\tabcolsep}{7pt}
\renewcommand{\arraystretch}{1.4}
\tiny
\caption{Link prediction results with AUC scores over 10 different runs.}
\vspace{-0.4cm}
\begin{center}
\begin{tabular}{|c|c|c|c|c|c|c|}
\hline
% \multirow{2}{*}{Dataset} &
Method & Bitcoin-Alpha & Bitcoin-OTC & Slashdot & WikiElection & WikiEditor & WikiRfA \\  
\hline 
SiNE & 0.8351 $\pm$ 0.0126 & 0.8575 $\pm$ 0.0053 & 0.8108 $\pm$ 0.0021 & 0.8040 $\pm$ 0.0072 & 0.8631 $\pm$ 0.0044 & 0.7963 $\pm$ 0.0260 \\
SLF & 0.8438 $\pm$ 0.0151 & 0.8670 $\pm$ 0.0052 & \textbf{0.8846 $\pm$ 0.0040} & 0.8803 $\pm$ 0.0025 & 0.9090 $\pm$ 0.0027 &  \underline{0.8709 $\pm$ 0.0012} \\
SGCN & 0.8420 $\pm$ 0.0147 & 0.8780 $\pm$ 0.0103 & 0.8543 $\pm$ 0.0064 & 0.8516 $\pm$ 0.0030 & 0.9068 $\pm$ 0.0040 & 0.8361 $\pm$ 0.0078 \\
SNEA & 0.8453 $\pm$ 0.0062 & 0.8792 $\pm$ 0.0028 & 0.8621 $\pm$ 0.0154 & 0.8412 $\pm$ 0.0083 & 0.9278 $\pm$ 0.0062 & 0.8259 $\pm$ 0.0036\\
\singh{SDGNN} & 0.9008 $\pm$ 0.0081 & 0.9128 $\pm$ 0.0073 & \underline{0.8734 $\pm$ 0.0187} & 0.8763 $\pm$ 0.0134 & 0.9430 $\pm$ 0.0126 & \textbf{0.8870 $\pm$ 0.0048} \\
Spectral-SGCN-I & 0.9005 $\pm$ 0.0345 & 0.9079 $\pm$ 0.0176 & 0.8345 $\pm$ 0.0121  & 0.8559 $\pm$ 0.0316 & 0.9447 $\pm$ 0.0194 & 0.8228 $\pm$ 0.0333 \\
Spectral-SGCN-II & \underline{0.9146 $\pm$ 0.0066} & \underline{0.9309 $\pm$ 0.0044} & 0.8677 $\pm$ 0.0090  & \underline{0.8840 $\pm$ 0.0018} & \textbf{0.9818 $\pm$ 0.0047} &  0.8420 $\pm$ 0.0026 \\
Spectral-S2GCN & 0.8670 $\pm$ 0.0176 & 0.8936 $\pm$ 0.0095  & 0.8273 $\pm$ 0.0291 & 0.8149 $\pm$ 0.0315 &  0.9375 $\pm$ 0.0121& 0.8242 $\pm$ 0.0189 \\
Signed-MagNet & \textbf{0.9227 $\pm$ 0.0097} & \textbf{0.9410 $\pm$ 0.0076}  & 0.8615 $\pm$ 0.0074 & \textbf{0.8881 $\pm$ 0.0026} & \underline{0.9567 $\pm$ 0.0144} & 0.8612 $\pm$ 0.0032 \\
\hline
\end{tabular}
\end{center}
\label{table:link_prediction_auc}

\end{table*}
%-------------------------------------------

%=====================================
\section{Connections to SGCN}
\label{sec:connection}
SGCN~\citep{DerMaTan18} in its design consider balanced and unbalanced node sets based on balance theory in feature aggregation process. The balanced node set for a target node $i$ is the set of nodes that have even number of negative links along a path connecting to $i$. An $\ell$-hop balanced set of nodes for target node $i$ is denoted as $B_i(\ell)$ and unbalanced set of nodes as $U_i(\ell)$. For example graph in Figure~\ref{fig:toy1}, $B_1(1) = \{2\}$ and $U_1(1) = \{ 3,4\}$. 

The node representations for these balanced and unbalanced sets are treated separately in feature aggregation process and are concatenated together to form final node embeddings. In the $\ell^{th}$ layer of the model, it reads
\begin{align*}
    \bh_i^{B(\ell)} \!&= \sigma \!\! \left( \!\! \bTheta^{B(\ell)} \!\! \left[ \sum_{j\in \cN_i^+}\!\!\! \frac{\bh_j^{B(\ell-1)}}{|\cN_i^+|},\!\!\! \sum_{k\in \cN_i^-}\!\!\! \frac{\bh_j^{U(\ell-1)}}{|\cN_i^-|}, \bh_i^{B(\ell - 1)} \right]  \right) \\
    \bh_i^{U(\ell)} \!&= \sigma \!\! \left( \!\! \bTheta^{U(\ell)} \!\! \left[ \sum_{j\in \cN_i^+}\!\!\! \frac{\bh_j^{U(\ell-1)}}{|\cN_i^+|},\!\!\! \sum_{k\in \cN_i^-}\!\!\! \frac{\bh_j^{B(\ell-1)}}{|\cN_i^-|}, \bh_i^{U(\ell - 1)} \right]  \right),
\end{align*}
where $\bTheta^{B(\ell)}$ and $\bTheta^{U(\ell)}$ are linear transformation parameters for balanced and unbalanced paths, respectively and the node representation at $\ell^{th}$ layer is the concatenation of the two embeddings $\bh^{(\ell)} =[\bh_i^{B(\ell)}, \bh_i^{U(\ell)}] $.

Instead of treating positive and negative neighbors separately, in our architecture of Spectral-SGCN-I we are aggregating them weighted by their signs and corresponding (absolute) degrees as can be seen from Equation~\eqref{eq:aggr_lowpass}. 
% \chen{the connection is not clean. maybe for some specific choice of parameters, the two algorithms (SGCN and Spectral-SGCN-I) would coincide?}

\section{Effect of Input Feature Size}
\begin{figure}[ht]
\centering
\begin{subfigure}[t]{0.45\textwidth}
\centering	
\includegraphics[scale=0.24]{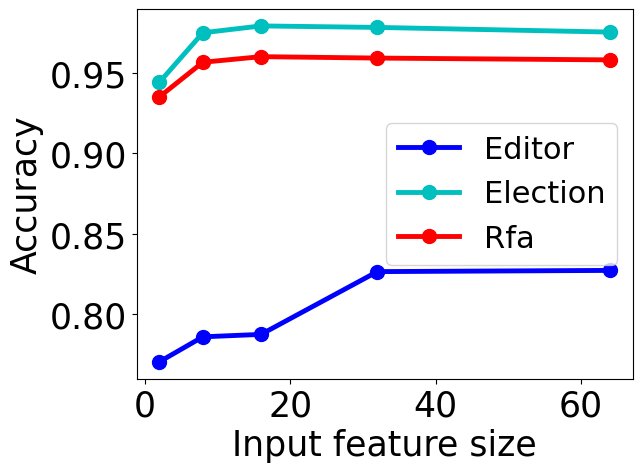}
\end{subfigure}
\begin{subfigure}[t]{0.45\textwidth}
\centering	
\includegraphics[scale=0.24]{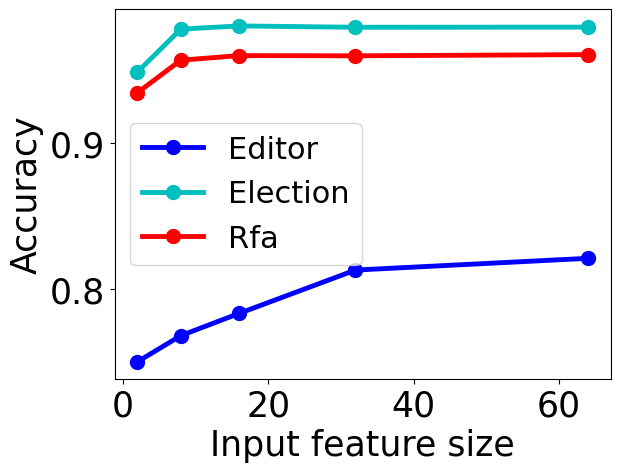}
\end{subfigure}
\caption{Spectral-SGCN-I (left) and Spectral-SGCN-II (right) performance with varying dimensions of input features (5\% known labels).}
\label{fig:vary_feat}
\end{figure}

We also perform the study on the classification accuracy with varying number of input feature sizes. Figure~\ref{fig:vary_feat} shows the performance of Spectral-SGCN-II with respect to varying number of input features. As expected, the performance improves with increase in the dimension of input features. 

%*************************
%*************************

% %+===================================
% \section{Rotation Property}

% With experimental results, we observe that when the direction of inter-cluster signs are reversed (starting from unidirectional inter-cluster directed negative signs), the embeddings of one cluster rotates clockwise keeping the embeddings of other cluster unchanged. 
% %*************************
% \begin{figure*}[ht]
% \centering
% \begin{subfigure}[t]{0.42\textwidth}
% \centering	
% \includegraphics[scale=0.26]{./figures/rotate_cl1_cl2.eps}
% \caption{Directed negative edges from green cluster to red cluster.}
% \end{subfigure} \hspace{0.5cm}
% \begin{subfigure}[t]{0.42\textwidth}
% \centering
% \includegraphics[scale=0.26]{./figures/rotate_cl2_cl1.eps}
% \caption{Directed negative edges from red cluster to green cluster.}
% \end{subfigure}
% \caption{Rotation results with 50 nodes per cluster.}
% \label{fig:rotation}
% \end{figure*}
% %****************************

\end{document}